\documentclass[11pt]{report}
\usepackage{algorithm}
\usepackage{algpseudocode}
\usepackage[nottoc]{tocbibind}
\usepackage{setspace}
\usepackage[frozencache=true,cachedir=minted-cache]{minted}
\usepackage{mdframed}
\surroundwithmdframed{minted}
\usepackage[utf8]{inputenc}
\usepackage{amsmath}
\usepackage{amsthm}
\usepackage{parskip}
\usepackage[export]{adjustbox}
\theoremstyle{definition}

\usepackage[dvipsnames]{xcolor}
\usepackage{amssymb}
\usepackage{tabularx}
\usepackage[justification=centering]{caption}
\usepackage[numbers]{natbib}

\BeforeBeginEnvironment{verbatim}{}
\usepackage{listings}
\usepackage{pgfplots}
\pgfplotsset{compat=1.9}
\usepackage{caption}
\usepackage{graphicx}
\usepackage{float}
\usepackage[margin=3cm]{geometry}
\usepackage{breqn}
\usepackage[english]{babel}
\usepackage{hyperref}
\addto\extrasenglish{%
}
\addto\extrasenglish{%
}
\hypersetup{
    colorlinks=true,
    linkcolor=blue,
    filecolor=blue,      
    urlcolor=blue,
    citecolor=blue,
    pdfpagemode=FullScreen,
    }

\graphicspath{ {./images/} }
\date{Trinity 2022}

\begin{document}
\begin{titlepage}
    \centering
    \includegraphics[width=4cm]{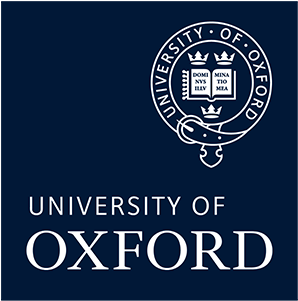}\par\vspace{1cm}
    {\scshape\Huge Constraint-driven multi-task learning \par}
    \vspace{1cm}
    {\scshape\Large Bogdan Cretu \par}
    {\scshape\Large University of Oxford \par}
    \vspace{5cm}

    \vfill
    Supervised by: Andrew Cropper \par
    {\large Trinity 2022 \par}
\end{titlepage}
\pagenumbering{arabic}
\begin{abstract}
\centering\begin{minipage}{\dimexpr\paperwidth-8cm}
\setlength{\parskip}{0.5\baselineskip plus 2pt}
Inductive logic programming is a form of machine learning based on mathematical logic that generates logic programs from given examples and background knowledge.

 In this project, we extend the Popper ILP system to make use of multi-task learning. We implement the state-of-the-art approach and several new strategies to improve search performance. Furthermore, we introduce constraint preservation, a technique that improves overall performance for all approaches. 
 
 Constraint preservation allows the system to transfer knowledge between updates on the background knowledge set. Consequently, we reduce the amount of repeated work performed by the system. Additionally, constraint preservation allows us to transition from the current state-of-the-art iterative deepening search approach to a more efficient breadth first search approach.
 
 Finally, we experiment with curriculum learning techniques and show their potential benefit to the field.
 \end{minipage}
\end{abstract}
\newpage
\addtocontents{toc}{\protect\thispagestyle{empty}}
\tableofcontents
\thispagestyle{empty}
\newpage
\chapter{Introduction}
\setcounter{page}{1}
\section{Motivation}
Recent Artificial Intelligence (AI) research has shown considerable interest in developing systems that can mimic human learning and thinking  \cite{motivationintro}. Similar to human learning, there has been significant success in transferring knowledge between AI systems, like a ``student" - ``teacher" relationship \cite{generalkt}. To simulate human learning, a ``student" system should also be able to accumulate knowledge and combine the obtained information. For example, it is not expected of a Computer Science pupil to learn how to implement quicksort without first understanding the notions of median and partition. We can extend this expectation to an AI system. Instead of solving a complex task without prior knowledge, firstly, learn how to resolve more straightforward problems from the same area. This form of Machine Learning is referred to as multi-task learning \cite{zhang2018overview}.

Inductive Logic Programming is an ideal candidate for the above use case \cite{ILPat30}.
Inductive Logic Programming (ILP) is a form of machine learning that learns logic programs (also referred to as symbolic programs) based on given examples and background knowledge (BK). The BK set consists of a set of logic programs given by the user. Therefore, transferring knowledge becomes trivial \cite{johnthesis}: if we want to reuse a solution in solving another task, it is sufficient to append it to the given BK. Returning to the ``student" - ``teacher" example, the previously mentioned approach would be equivalent to having a teacher provide his student with a new definition.

Aside from providing support for multi-task learning, ILP has been shown to provide extensive benefits compared to most ML models, such as Data Efficiency (being able to generalise from a small number of examples) and Explainability (providing outputs that are easily understandable by a human) \cite{ILPat30}. 

We will proceed by providing the reader with an example logic program (see \autoref{fig:ILPexample}) written in Prolog in order to accommodate them with the syntax used throughout the project. Afterwards, we will describe four problems that can be solved by an ILP system and which would benefit from multi-task learning.
\begin{figure}[hbt!]
    \begin{minted}{prolog}
    % Background knowledge:
        isFather("John", "Dan") % John is Dan's father
        isFather("Paul", "John") % Paul is John's father
        isWife("Alice", "Paul") % Alice is Paul's wife
        
        isGrandmother(X,Y):-isWife(X,Z),
                            isGrandfather(Z,Y) % X is Y's grandmother 
                                               % if X is Z's wife and
                                               % Z is Y's grandfather
                                                
        isGrandfather(X,Y):-isFather(X,Z),
                            isFather(Z,Y)  % X is Y's grandfather
                                           % if X is Z's father and
                                           % Z is Y's father
                                            
    % Logic programming query:
        ?-isGrandmother("Alice","Dan") % Is Alice Dan's grandmother?
    \end{minted}
    \caption{Logic program example}
    \label{fig:ILPexample}
\end{figure}

\paragraph{Kinship relationships.} Kinship relationships allow an illustration of the problem solved by introducing multi-task learning in ILP systems. 
Moreover, the trivial nature of the problem allows for synthetic data generation to benchmark. In \autoref{fig:kinship_example} we show a situation where a shorter program (\texttt{isGrandmother(A,B) :- isWife(A,C), isGrandfather(C,B)} instead of \texttt{isGrandmother(A,B) :- isWife(A,C), isFather(C, D), isFather(D,B)}) is learnt to explain the grandmother relationship. 
\begin{figure}[hbt!]
    \centering
    \begin{minted}{prolog}
    % Background knowledge:
        isFather("John","Dan")
        isFather("Paul","John")

        isWife("Alice","Paul")
        
    % Given examples:
        isGrandfather("Paul","Dan")
        
        isGrandmother("Alice","Dan")
        
    % Programs learnt:
        isGrandfather(A,B) :- isFather(A,C), isFather(C,B)
        
        isGrandmother(A,B) :- isWife(A,C), isGrandfather(C,B)
        
    % Programs learnt without multi task learning:
        isGrandfather(A,B) :- isFather(A,C), isFather(C,B)
        
        isGrandmother(A,B) :- isWife(A,C), isFather(C, D), isFather(D,B)
    \end{minted}
    \caption{Example for the kinship relationship problem}
    \label{fig:kinship_example}
\end{figure} 

\paragraph{String transformation.} One of the most widely used examples to benchmark ILP systems are EXCEL type string transformation operations \cite{ILPat30}. 
Through multi-task learning, an ILP system can better capture situations where a complex transformation consists of sequential applications of more simple operations. In \autoref{fig:string_example}, we present our initial set of example and their equivalent programs. The given functions perform the following operations:
\begin{itemize}
    \item \texttt{f0} - capitalize each word of a string
    \item \texttt{f1} - capitalize a single word
    \item \texttt{f2} - drop the first word of a string and capitalize the second
    \item \texttt{f3} - copy a string until the first non-letter character is encountered
\end{itemize}
The predicates \texttt{inv0} and \texttt{inv2} from the example are auxiliary functions invented during the learning process.

\begin{figure}[hbt!]
    \centering
    \begin{minted}{prolog}
    % Given examples:
        f0('gerson zaverucha', 'Gerson Zaverucha')
        f0('jesse davis', 'Jesse Davis')
        f0('paolo frasconi', 'Paolo Frasconi')
        
        f1('james', 'James')
        f1('jozie', 'Jozie')
        f1('paul', 'Paul')
        
        f2('dr wilson', 'Wilson')
        f2('dr brown', 'Brown')
        f2('dr wright', 'Wright')
        
        f3('artificial.', 'artificial')
        f3('intelligence ', 'intelligence')
        f3('systems1a', 'systems')

    % Programs learnt:
        f0(A,B):-f1(A,C),inv0(C,B).
        inv0(A,B):-copyskip1(A,C),f1(C,B).
        
        f1(A,B):-mk_uppercase(A,C),f3(C,B).
        
        f2(A,B):-skip1(A,C),inv2(C,B).
        inv2(A,B):-skip1(A,C),f1(C,B).
        inv2(A,B):-skip1(A,C),inv2(C,B).
        
        f3(A,B):-copyskip1(A,B),not_letter(B).
        f3(A,B):-copyskip1(A,C),f3(C,B).
    \end{minted}
    \caption{Example for the string transformation problem}
    \label{fig:string_example}
\end{figure} 
\paragraph{Robot movement.} Robot movement can be represented as basic movements such as going up/down/left/right on a grid. For traversing large distances or completing complex paths it would be more efficient to have common patterns such as corner turns, as part of the background knowledge. 

In \autoref{fig:robot_example}, we show a solution where a 5 step translation is reduced to only two literals.
\begin{figure}[hbt!]
    \centering
    \begin{minted}{prolog}
    % Data format:
        f(State(X, Y, Xnew, Ynew, Width, Height))
        
    % Given examples:
        turn(State(0, 1, 1, 2, 5, 2))
        turn(State(0, 0, 1, 1, 5, 2))
        
        jump(State(0, 1, 1, 4, 5, 2))
        jump(State(0, 0, 0, 3, 5, 2))
        
        turnAndJump(State(0, 0, 1, 4, 5, 2))

    % Programs learnt:
        turn(State(X, Y, Xnew, Ynew, W, H)) :- right(X, Y, X2, Y2),
                                               down(X2, Y2, Xnew, Ynew)
                                                    
        jump(State(X, Y, Xnew, Ynew, W, H)) :- right(X, Y, X2, Y2),
                                               right(X2, Y2, X3, Y3)
                                               right(X3, Y3, Xnew, Ynew)
                                                        
        turnAndJump(State(X, Y, Xnew, Ynew, W, H)) :- 
                                                turn(X, Y, X2, Y2),
                                                jump(X2, Y2, Xnew, Ynew)
    \end{minted}
    \caption{Example for the robot movement problem}
    \label{fig:robot_example}
\end{figure} 

\paragraph{Drawing images.} Most graphic rendering techniques are performed by decomposing scenes into primitives such as lines, triangles or quads. Specific algorithms (such as Constructive Solid Geometry) also make use of more complex shapes, for which the primitive decomposition is computed in advance \cite{Graphics21}.

Based on the above, we formulate the following problem: given a set of functions that control a printer head (moving over the four axes, returning to (1,1) coordinates or drawing a pixel) the system has to learn to draw complex images. In \autoref{fig:drawing-images} we show how a cube can be drawn by connecting four lines.
\begin{figure}[hbt!]
    \centering

    \begin{minted}{prolog}
    % Data format:
        f(State(X, Y, Width, Height, Grid))
        
    % Given examples:
        topLine(State(_, _, 3, 3, [1,1,1,0,0,0,0,0,0]))
        bottomLine(State(_, _, 3, 3, [0,0,0,0,0,0,1,1,1]))
        leftLine(State(_, _, 3, 3, [1,0,0,1,0,0,1,0,0]))
        rightLine(State(_, _, 3, 3, [0,0,1,0,0,1,0,0,1]))
        isCube(State(_, _, 3, 3, [1,1,1,1,0,1,1,1,1]))
        
    % Programs learnt:
        topLine(S, S5) :- at_start(S), move_up(S, S2), move_up(S, S3), 
                             draw1(S3), move_right(S3, S4), draw1(S4), 
                             move_right(S4, S5), draw1(S5)
        bottomLine(S, S3) :- at_start(S), draw1(S), move_right(S, S2), 
                                draw1(S2), move_right(S2, S3) draw1(S3)
        leftLine(S, S3) :- at_start(S), draw1(S), move_up(S, S2), 
                              draw1(S2), move_up(S2, S3) draw1(S3)
        rightLine(S, S5) :- at_start(S), move_right(S, S2), 
                               move_right(S, S3), draw1(S3), 
                               move_up(S3, S4), draw1(S4), 
                               move_up(S4, S5), draw1(S5)
        isCube(S, S4) :- topLine(S, S2), returnStart(S2, S3), 
                         bottomLine(S3, S4), returnStart(S4, S5),
                         leftLine(S5, S6), returnStart(S6, S7),
                         rightLine(S7, S8), returnStart(S8, S9)
    \end{minted}
    \caption{Example for the image drawing problem}
    \label{fig:drawing-images}
\end{figure}

All the above problems are solvable by an ILP system, but certain limitations exist. ILP systems learn by searching through a hypothesis space of possible solutions, but even the most advanced search techniques face difficulties when the size of the hypothesis space increases significantly. The size of the hypothesis space is directly related to the maximum size a program can have \cite{popper}. In order to better understand this statement, imagine the task of writing down all possible programs of size $n$ that contain a given list of predicates (allowing duplicates). One can easily observe that the size of the result grows exponentially with $n$. Therefore, the maximum size of possible solutions must be restricted such that searching through it is feasible. Consequently, if the solution for a task is too large, the task will become unsolvable. 

Multi-task learning mitigates the above issue by reusing learnt solutions when learning tasks from related domains. In the four previous examples, we can observe that several tasks can be represented as more compact solutions by reusing the solutions from some other tasks.

The state-of-the-art approach to ILP multi-task learning is based on the Metagol \cite{metagol} system and was described by \citet{lin2014bias} in 2014. It has been shown to approach the performance of commercial systems of the time and outperform an independent learning strategy (running Metagol on each task separately). In \autoref{implementation} we will identify significant drawbacks of the current strategy that negatively impact its learning abilities. 

Finally, we introduce Popper \cite{popper}, an ILP system which has superseded Metagol \cite{metagol}. Popper uses a new approach to ILP, referred to as ``Learn from failures": while searching through the hypothesis space, it generates constraints based on incorrect programs found. Those constraints are then used to prune the search space and improve learning time. Furthermore, in \autoref{constraints} we detail how constraints can be further used to improve the learning rate for multi-task systems.

\section{Contribution} 

Our contribution to the field represents an enhanced version of Popper that supports multi-task learning and uses knowledge transfer. 

We have decided to implement our approach and re-implement the method described by \citet{lin2014bias} as extensions to the Popper ILP system in order to be able to evaluate the benefit that multi-task learning can obtain by preserving constraints.

Such, we summarise our contribution as follows:
\begin{itemize}
    \item Metagol$_{DF}$ performs iterative deepening search based on the maximum size of programs in the hypothesis space. We identify situations where increasing search size is detrimental to the learning process, virtually blocking the system from progressing further. In consequence, we provide an alternative search strategy that circumvents this obstacle and has a greatly improved learning ability.
    \item Between each stage of iterative deepening search, the Metagol$_{DF}$ does not maintain any information. Our implementation gives Popper the ability to preserve constraints when the maximum search size is modified. Such, we improve the learning rate and allow for a breadth first search approach to replace the iterative deepening approach. For example, let $C$ be the set of constraints obtained by searching through a hypothesis space of size $n$. This constraints will help prune a search space of size $n+1$. Without constraint preservation, the system would be unable to transfer information between runs. Therefore, it must search through a space size that includes both hypotheses of size $n$ and size $n+1$.
    \item We show that the performance of the system is strongly correlated to the order in which the user inputs the list of tasks to be solved \cite{ordering}. We design two methods that order tasks based on how close the system is to solving them, aiming to achieve more consistent and closer to optimal performance.
    \item Metagol$_{DF}$ has been tested against a limited data set of 17 string transformation operations. We refine one external data set and synthetically generate two additional ones. We then proceed to empirically evaluate the performance of all proposed techniques and indicate how each performs in comparison to the original approach.
\end{itemize}

\chapter{Related work}

\section{Ability to learn}
Programs that expand their learning abilities (``have the ability to learn") have been an area of interest in Machine Learning for multiple decades. In \cite{Solomonoff1989ASF}, Solomonoff describes a system that, given a small set of concepts, is able to learn problems of increasing difficulty by extracting new, more complex concepts during each learning stage. A concept was defined as a sequence of instructions. The approach reduces most problems from science to two main categories: 

\textbf{Inversion problems.} Given a machine $M$, that performs string transformation operations, and a string $s$, the goal of an inversion problem is to determine a string $x$, such as $M(x) = s$. 

\textbf{Time limited optimisation problems.} Given a machine $M$, that takes a string and returns a real number, the goal of a time limited optimisation problem is to find a string $x$, within a fixed time limit $T$, such as $M(x)$ is as large as possible. 

Furthermore, the paper concludes with an overview of areas where such a system would benefit from further research, most notably: improvements in the method used to update the set of knowledge once a new task is learnt and designing a system that is able to work on an unordered set of programs.

\section{Multi-task learning}
In this thesis, we refer to multiple Machine Learning paradigms related to multi-task learning. 

\textbf{Meta-learning.} Meta-learning (learning to learn) focuses on how machine learning approaches perform on certain tasks and then uses information obtained from previous runs in order to speed up learning during future tasks \cite{Joaquin2018}. 

\textbf{Curriculum learning.} Curriculum learning means that tasks will be learnt in a specific order, based on their perceived difficulty (similar to a human curriculum) \cite{XinWang2020}. 

\textbf{Transfer learning.} Transfer learning is a technique through which multiple learners (that focus on related tasks) transfer knowledge to each other.  \cite{weiss2016survey}. 

The above approaches have been empirically and theoretically shown to improve performance over traditional independent learning methods \cite{zhang2018overview}. 

Although the fact that tasks from a similar domain may be related appears intuitive, it is necessary to outline both the benefits and downsides of multi-task learning. Therefore, depending on the similarity between the given set of tasks, it is crucial to evaluate if multi-task learning is expected to provide a guaranteed advantage over independent learning \cite{ben2003exploiting}.
The expectation is that multi-task learning approaches perform significantly worse than conventional learning methods on unrelated tasks \cite{pan2010}. Moreover, transfer learning systems that retain information from all learnt tasks as part of their knowledge set face difficulty when used to solve large data sets \cite{ferri2001incremental}. This is a consequence of the fact that search speed is usually correlated to the size of the background knowledge.
Several methods that attempt to circumvent the above issues have been developed. 

For example, neural networks have been used to recognise which programs are expected to perform better on specific input/output pairs \cite{ellis2018learning}. 

It has also been experimented with the idea of ``forgetting", which refers to removing data from the background knowledge (or ``unlearning") with the purpose of improving search time \cite{cropper2020forgetting}.
\section{Multi-task learning in ILP}
One of the main advantages of ILP, in the context of multi-task learning, is the fact that ILP systems can perform knowledge transfer without significant overhead through the fact that outputs are reusable as inputs. Despite such considerations, very few studies explore multi-task learning approaches for ILP. 

Metagol is an ILP system based on meta-interpretive learning \cite{metagol}. Lin et al. adapt the Metagol system to make use of multi-task learning (Metagol$_{DF}$) and benchmark it against multiple agents: a human agent, a commercial system (Microsoft FlashFill, introduced with Excel 2013), and a version of itself, which does not make use of multi-task learning. 

The system has shown promising results and consistently achieved better or equal performance when compared with the standard approach. 

The author concludes by highlighting a series of limitations of multi-task learning in the context of ILP.
Most importantly, it observes that the system will become overwhelmed when run on multiple data sets. This claim is empirically confirmed by Cropper, in \cite{cropper2020forgetting}, who also provides a solution for such use cases: introducing Forgetgol. This system can forget programs and reduce the background knowledge size. 

In this project, we identify significant issues with the previous approach and propose an alternative solution. We build on the Popper ILP system \cite{Cropper2019} and introduce the following elements of novelty: we modify how the hypothesis space is searched, we retain information between failed search attempts, and we introduce new search strategies that focus on analysing the difficulty of the tasks at run-time (similar to curriculum learning).

\chapter{Problem setting}
In this chapter, we will give an overview of ILP in order to familiarise the reader with specific terminology used throughout the project. 
\section{Logic programming syntax}
We will define the syntax of Logic Programs by following definitions from \cite{ILPat30}, and \cite{kowalski1974}: 

\textbf{Variable.} A variable is a string of characters that starts with an uppercase letter. 

\textbf{Function.} A function symbol is a string of characters that starts with a lowercase letter. Examples: \texttt{f1}, \texttt{move\_up}, \texttt{draw1}. 

\textbf{Predicate symbol.} A predicate symbol is a string of characters that start with a lowercase letter, similarly to a function. 

\textbf{Arity.} The arity of a function/predicate symbol \texttt{f} is the number of arguments it takes and is denoted as \texttt{f/n}. 

\textbf{Constant symbol} A constant symbol is a predicate symbol of arity zero. Examples: \texttt{alice}, \texttt{bob}. 

\textbf{Term.} A term refers to either a variable, a constant symbol or a predicate symbol of arity $n$ followed by a tuple of size $n$. 

\textbf{Ground term.} We say that a term is ground if it contains no variables. 

\textbf{Atom.} An atom is a formula $f(x_1,..,x_n)$, such as $p$ is a predicate symbol of arity $n$ and $x_i$ is a term $\forall i \in \{1,..,n\}$. 

\textbf{Ground atom.} We say that an atom is ground if all its terms are ground. 

\textbf{Negation.} We say that an atom is false if it cannot be proven true. Throughout this paper, we will represent negation through the symbol $\neg$. 

\textbf{Literal.} A literal is either an atom or its negation. 

\textbf{Clause.} A clause is a disjunction of atoms $h_1,h_2,..,h_n,b_1,b_2,..,b_m$, represented as:
\begin{align*}
    h_1,h_2,..,h_n \textrm{ :- } b_1,b_2,..,b_m
\end{align*}
The above definition is equivalent to:
\begin{align*}
    (h_1\land h_2\land..\land h_n) \lor \neg b_1 \lor \neg b_2 .. \lor \neg b_m
\end{align*}

We refer to $h_1,h_2,..,h_n$ as the \textit{head literals} and  $b_1,b_2,..,b_m$ as the \textit{body literals} of the clause. 

The above representation is shorthand for ``$h$ holds true if all of $b_1,..,b_m$" hold true. 

\textbf{Horn clause} A Horn clause is a clause for which $n$ (number of head/positive literals) is at most $1$. 

We will only be concerned with Horn clauses since other types of clauses are not supported by basic Prolog \cite{kowalski1988early}. 

\textbf{Ground clause.} We say that a clause is ground if it contains no variables. 

\textbf{Definite clause.} We say that a Horn clause is definite if the number of positive literals is exactly one. 

\textbf{Definite logic program.} A definite logic program is a set of definite clauses. We will further refer to definite logic programs as programs.
\section{Inductive logic programming}
Inductive logic programming problems are given as triplets $(B,E^+,E^-)$, where $B$ represents the background knowledge (BK), $E^+$ represents the set of positive examples, and $E^-$ represents the set of negative examples. 

\textbf{Background knowledge.} The background knowledge is a set of programs for which the definition is given completely. By the close world assumption, anything that is not explicitly stated true in the BK is considered false \cite{REITER1981119}. We show a specimen from the BK of the string transformation problem: 

\[
\texttt{B} = \left\{
\begin{array}{l}
    \texttt{is\char`_empty([]).}\\
    \texttt{is\char`_space([' '|\char`_]).}\\
    \texttt{is\char`_number(['0'|\char`_]).}\\
    \texttt{...} \\
    \texttt{is\char`_number(['9'|\char`_]).}\\
\end{array}
\right\}
\]

\textbf{Examples.} Examples are given as two sets (positive and negative examples) and are represented similarly to the BK, but all clauses must be ground. Moreover, there is no assumption on the function behaviour outside of the given use cases. Below we show examples of a function that turns the first letter of a string to uppercase. 

\[
\texttt{E$^+$} = \left\{
\begin{array}{l}
\texttt{f(['j', 'a', 'm', 'e', 's'],['J', 'a', 'm', 'e', 's']).}\\
\texttt{f(['j', 'o', 'z', 'i', 'e'],['J', 'o', 'z', 'i', 'e']).}\\
\texttt{f(['b', 'e', 'n'],['B', 'e', 'n']).}\\
\end{array}
\right\} 
\]

\[
\texttt{E$^-$} = \left\{
\begin{array}{l}
\texttt{f(['j', 'a', 'm', 'e', 's'],['j', 'a', 'm', 'e', 's']).}\\
\texttt{f(['j', 'o', 'z', 'i', 'e'],['o', 'z', 'i', 'e']).}\\
\texttt{f(['b', 'e', 'n'],['B']).}\\
\end{array}
\right\}
\]

\textbf{Hypothesis space.} The hypothesis space contains all the possible programs that can be built from the predicates given in the BK. Since the hypothesis space is infinite, it is important to enforce certain restrictions to make searching through it practical. Such, many machine learning makes use of inductive bias and language bias \cite[p,64,106]{Mitchell97} to restrict the hypothesis space. 
\section{Popper ILP System}
In \autoref{implementation} we introduce our multi-task learning strategies, which build on the Popper ILP system. The Popper ILP system uses an approach known as ``Learning from failures" \cite{popper}. This technique divides the learning process into three stages: \textbf{generate}, \textbf{test}, and \textbf{constrain}. 

In the generate phase, a logic program (referred to as hypothesis) is selected, such that it respects the currently learnt \textbf{hypothesis constraints}. Afterwards, the program is tested against the given examples. If all examples are covered, Popper will return the program as a valid answer. Otherwise, we proceed to the constrain stage, where information from the testing stage is used in order to augment the set of \textbf{hypothesis constraints}. Those three stages are repeated until either a solution is found or there are no more programs in the hypothesis space. 

Below we will describe the four types of constraints. 

\textbf{Generalisation constraints.} A generalisation constraint is a constraint that eliminates generalisations of the current hypothesis. Example: 

Given the following negative examples:
\[
\texttt{E$^-$} = \left\{
\begin{array}{l}
\texttt{f(['j', 'a', 'm', 'e', 's'],['J', 'a', 'm', 'e', 's']).}\\
\texttt{f(['b', 'e', 'n'],['B', 'e', 'n']).}\\
\end{array}
\right\} 
\]

And the hypothesis:
\[
\texttt{h} = \left\{
\begin{array}{l}
\texttt{f(A, B):-mk\_uppercase(A,B).}\\
\end{array}
\right\} 
\]

We observe that a negative example holds for $h$. Such, we can prune generalisations, such as:
\[
\texttt{h'} = \left\{
\begin{array}{l}
\texttt{f(A, B):-mk\_uppercase(A,B).}\\
\texttt{f(A, B):-head(A,B).}\\
\end{array}
\right\} 
\]

\textbf{Specialisation constraints.} A specialisation constraint is a constraint that prunes out specialisations of the current hypothesis. Example: 

Given the following positive examples:
\[
\texttt{E$^-$} = \left\{
\begin{array}{l}
\texttt{f(['j', 'a', 'm', 'e', 's'],['J', 'a', 'm', 'e', 's']).} \\
\texttt{f(['b', 'e', 'n',' ', 'g'],['B', 'e', 'n',' ', 'G']).} \\
\end{array}
\right\} 
\]

And the hypothesis:
\[
\texttt{h} = \left\{
\begin{array}{l}
\texttt{f(A, B):-mk\_uppercase(A,B).}\\
\end{array}
\right\} 
\]

We observe that the first positive example holds for $h$, but not the second. Such, we conclude that $h$ is too specific and prune out specialisations of $h$, such as:
\[
\texttt{h'} = \left\{
\begin{array}{l}
\texttt{f(A, B):-mk\_uppercase(A,B),empty(B).}\\
\end{array}
\right\} 
\]

\textbf{Elimination constraints.} An elimination constraint is a constraint that eliminates a hypothesis $h$ from any separable hypothesis $h_s$ that contains $h$. We refer to a hypothesis $h_s$ as separable if no predicate symbol from the head of a clause in $h_s$ appears in the body of a clause in $h_s$. Example: 

Given the following positive examples:
\[
\texttt{E$^-$} = \left\{
\begin{array}{l}
\texttt{f(['j', 'a', 'm', 'e', 's'],['J', 'a', 'm', 'e', 's']).} \\
\texttt{f(['b', 'e', 'n',' ', 'g'],['B', 'e', 'n',' ', 'G']).} \\
\end{array}
\right\} 
\]

And the hypothesis:
\[
\texttt{h} = \left\{
\begin{array}{l}
\texttt{f(A, B):-empty(A).}\\
\end{array}
\right\} 
\]

We observe that the no positive example holds for $h$. Such, we conclude that $h$ can be eliminated from any separable hypothesis, such as:
\[
\texttt{h'} = \left\{
\begin{array}{l}
\texttt{f(A, B):-mk\_uppercase(A,B)}\\
\texttt{f(A, B):-empty(A).}\\
\end{array}
\right\} 
\]

\textbf{Banish constraints.} A banish constraint is a constraint that rules out one specific hypothesis. 

To handle searching through a vast hypothesis space size, the Popper ILP system looks for solutions in increasing order of the number of literals in a hypothesis. The formula used to compute an upper bound on the size of the hypothesis space is given in \cite{popper}: 
\begin{itemize}
    \item Maximum arity $\boldsymbol{a}$.
    \item Maximum number of unique variables in a clause $\boldsymbol{v}$.
    \item Maximum number of body literals allowed in a clause $\boldsymbol{m}$.
    \item Maximum number of clauses in a hypothesis $\boldsymbol{n}$.
    \item Declaration bias $\boldsymbol{D}=(D_h,D_b)$.
\end{itemize}

\begin{align*}
    \sum_{j=1}^{n}\binom{|D_h| v^a \sum_{i=1}^{m}\binom{|D_b| v^a}{i}}{j}
\end{align*}
\label{ILPintro}
\chapter{Implementation}
\label{implementation}
In this section, we will describe several multi-task learning approaches. For each one, we will detail the advantages and disadvantages. We begin with the naive approach, followed by the current state-of-the-art strategy. Afterwards, we will provide a new search algorithm that focuses on resetting the size of the search space in order to overcome the limitations of the state-of-the-art approach. Finally, we will describe two refinements that can be applied to any of the described approaches: constraint preservation and automatic task ordering.

\section{Naive approach}
\noindent This naive approach to multi-task learning iterates through all the tasks in order and attempts to learn them by calling an ILP system (in our case Popper) \footnote{We note that this search strategy can get blocked when working on unsolvable tasks. This behaviour does not impact our results as the naive approach is only used for reference and we will benchmark against the state-of-the-art strategy (which is discussed in \autoref{sota-sub}). Other approaches that do not make use of multi-task learning can be designed to overcome this difficulty.}. Therefore, it does not make use of the fact that tasks learned together are expected to be correlated. 

\paragraph{Advantages.} The main advantage of this approach is that it does not suffer from the additional overhead introduced by increasing the number of tasks in the BK \cite{lin2014bias}. In \autoref{fig:naive-example} the system is given two independent tasks: learning the \texttt{isGrandfather} and \texttt{isGrandmother} functions. In such a situation, transferring knowledge between the two will be redundant and only cause unnecessary overhead. Therefore, such use-cases benefit the naive approach.

\begin{figure}[hbt!]
    \centering

    \begin{minted}{prolog}
    
    % Background knowledge:
        isFather("John","Dan")
        isFather("Paul","John")

        isMother("Alice","Dan")
        isMother("Ana", "Alice")
        
    % Given examples:
        isGrandfather("Paul","Dan")
        
        isGrandmother("Ana","Dan")
        
    % Programs learnt:
        isGrandfather(A,B) :- isFather(A,C), isFather(C,B)
        
        isGrandmother(A,B) :- isMother(A,C), isMother(C, B)
        
    \end{minted}
    \caption{Advantage example for naive approach}
    \label{fig:naive-example}
\end{figure}

\paragraph{Disadvantages.} Although it performs well on uncorrelated data, multi-task learning systems have been shown repeatedly to out-perform traditional systems on tasks from the same domain \cite{ILPat30}. In \autoref{fig:bad-naive-example} the system is given the same two tasks, but the BK is modified such that the tasks are no longer independent: 

\texttt{isGrandmother} can be learnt as a function of \texttt{isGrandfather}. The naive approach has no provision to benefit from such situations.

\begin{figure}[hbt!]
    \centering

    \begin{minted}{prolog}
    
    % Background knowledge:
        isFather("John","Dan")
        isFather("Paul","John")

        isWife("Alice","Paul")
        
    % Given examples:
        isGrandfather("Paul","Dan")
        
        isGrandmother("Alice","Dan")
        
    % Programs learnt:
        isGrandfather(A,B) :- isFather(A,C), isFather(C,B)
        
        isGrandmother(A,B) :- isWife(A,C), isGrandfather(C,B)
        
    % Programs learnt without multi task learning:
        isGrandfather(A,B) :- isFather(A,C), isFather(C,B)
        
        isGrandmother(A,B) :- isWife(A,C), isFather(C, D), isFather(D,B)
        
    \end{minted}
    \caption{Disadvantage example for naive approach}
    \label{fig:bad-naive-example}
\end{figure}

Finally, in \autoref{alg:multi_popper_naive} we show an implementation of the naive multi-task learning approach.

\begin{algorithm}[hbt!]
\caption{Naive approach to multi-task learning through Popper}\label{alg:multi_popper_naive}
\begin{algorithmic}
\Function{multipoppernaive}{}
\State $T \gets \{t_1, t_2,..,t_n\}$
\State $\mathit{BK} \gets \{$BACKGROUND KNOWLEDGE$\}$
\ForAll{$t$ in $T$}
\State $S$ $\gets$ \{\}
\State $\mathit{size} \gets 1$
\While{TRUE}
    \State $\mathit{sol} \gets$ \Call{popper}{$t$, $BK$, $\mathit{size}$}
    \If{$\mathit{sol} \neq \mathit{null}$}
    \State $S$ $\gets$ $S$ $\cup \{\mathit{sol}\}$
    \State $T$ $\gets$ $T$ $\setminus$ \{$t$\} 
    \State \textbf{break}
    \EndIf
\State $\mathit{size} \gets \mathit{size}+1$ 
\EndWhile
\EndFor
\State \Return $S$
\EndFunction
\end{algorithmic}
\end{algorithm}

\section{Knowledge transfer: iterative deepening approach}
\label{sota-sub}
As previously discussed, the naive approach does not make use of knowledge transfer. To address this issue, in \cite{lin2014bias}, Lin et al. propose a new multi-task learning approach based on the Metagol ILP system. This method is currently state-of-the-art in multi-task ILP. The described solution uses iterative deepening search, proceeding as follows:
\begin{enumerate}
    \item Set $\mathit{maxSize} = 1$.
    \item For all tasks, try to find a solution with size at most $\mathit{maxSize}$.
    \item If all tasks are solved, return.
    \item Otherwise, update the background knowledge with newly learnt tasks, increment $\mathit{maxSize}$ by 1, and return at step 2.
\end{enumerate}

Pseudo-code for integrating this search strategy into the Popper ILP system is present in \autoref{alg:multi_popper_soda}. 

\begin{algorithm}[hbt!]
\caption{Iterative deepening search}\label{alg:multi_popper_soda}
\begin{algorithmic}
\Function{multipopper}{}
\State $T \gets \{t_1, t_2,..,t_n\}$
\State $\mathit{BK} \gets \{$BACKGROUND KNOWLEDGE$\}$
\State $S$ $\gets$ \{\}
\State $\mathit{maxSize} \gets 1$
\While{TRUE}
\For{$t$ in $T$}
    \State $\mathit{sol} \gets$ \Call{popper}{$t$, $\mathit{BK}$, $\mathit{maxSize}$}
    \If{$\mathit{sol} \neq \mathit{null}$}
    \State $S$ $\gets$ $S$ $\cup \{\mathit{sol}\}$
    \State $T$ $\gets$ $T$ $\setminus$ \{$t$\} 
    \EndIf
\EndFor
\State $\mathit{BK} \gets \mathit{BK} \cup S$
\If{$T =$ \{\}}
\State \Return $S$
\EndIf
\State $\mathit{maxSize} \gets \mathit{maxSize} + 1$
\EndWhile
\EndFunction
\end{algorithmic}
\end{algorithm}

\paragraph{Advantages.} Through the introduction of knowledge transfer as part of the multi-task learning strategy, this solution improves performance over the naive approach when used on correlated data sets. For example, assume our background knowledge contains two primitive predicates \texttt{bk0} and \texttt{bk1} and we are given two tasks, \texttt{f0} and \texttt{f1}, such that \linebreak \texttt{f1(A,B,C):-f0(A),f0(B),f0(C)}. Learning \texttt{f0} and re-using it's definition as part of an enhanced background knowledge allows us to learn \texttt{f1} more quickly by reducing the size of its solution.
\paragraph{Disadvantages.} We observe that even when the BK is augmented, the search size continues to increase. On one hand, this particularity provides an advantage when solution sizes remain large even with the new BK, such as in \autoref{ref:soda-example}. On the other hand, it gives rise to the following disadvantages: 

\begin{figure}[hbt!]
\begin{minted}{prolog}
    % Background knowledge:
        isFather("John","Dan")
        isFather("Paul","John")

        isWife("Alice","Paul")
        isMother("Mary","Alice")
        
    % Given examples:
        isGrandfather("Paul","Dan")
        
        isGrandgrandmother("Mary","Dan")
        
    % Programs learnt:
        isGrandfather(A,B) :- isFather(A,C), isFather(C,B)
        
        isGrandgrandmother(A,B) :- isMother(A,C), isWife(C,D),
                                   isGrandfather(D, B)
\end{minted}
\caption{Positive example for iterative deepening search}
\label{ref:soda-example}
\end{figure}

Firstly, after each increase of the $\mathit{maxSize}$ variable, we will reevaluate all programs of size between $1$ and  $\mathit{maxSize} - 1$ from the previous iteration. 

Secondly, we might search through an unnecessarily large hypothesis space when a BK augmentation might have allowed us to find a solution in a smaller space. 

Moreover, as a consequence of the second disadvantage, this approach does not learn the most compact solution possible. We will further describe this case below:
\begin{enumerate}
    \item We will use the BK from the robot movement problem. For simplicity, we will only be concerned with \texttt{move\_up(A,B)}.
    \item Assume we have two tasks, \texttt{f0(A,B)} and \texttt{f1(A,B)}, such that their solutions are:
    \begin{itemize}
        \item \texttt{f0(A,B):-move\_up(A,C),move\_up(C,D),move\_up(D,B)}
        \item \texttt{f1(A,B):-move\_up(A,C),move\_up(C,D),move\_up(D,E),move\_up(E,B)}
    \end{itemize}
    \item In this approach, we will find the solution for \texttt{f0} after having $\mathit{maxSize}=4$, and then increment $\mathit{maxSize}$ to 5, before augmenting the BK.
    \item When searching for programs up to size 5, we can identify two valid solutions for \texttt{f1}:
    \begin{itemize}
        \item  \texttt{f1(A,B):-move\_up(A,C),move\_up(C,D),move\_up(D,E),move\_up(E,B)}
        \item  \texttt{f1(A,B):-f0(A,C),move\_up(C,B)}
    \end{itemize}
    \item Such, we observe that the algorithm could return either solution, having no preference for the more compact program.
\end{enumerate}

Thirdly, this approach may lead to situations where learning does not conclude in a reasonable time frame because of the increase in the hypothesis space size. This can be encountered in a situation where functions a learnt in a chain, for example:
\begin{enumerate}
    \item Define $\texttt{p(A,B)}^n \texttt{:-} \texttt{p(A,C),p(C,B)}^{n-1}$ for any predicate \texttt{p} and natural number $n > 1$. Base case $\texttt{p(A,B)}^1 \texttt{:-} \texttt{p(A,B)}$. Then, we have tasks \texttt{f0,f1,f2,f3,f4}:
    \begin{itemize}
        \item $\texttt{f0(A,B):-move\_up(A,C)}^6$
        \item $\texttt{f1(A,B):-move\_up(A,C)}^{36}$
        \item $\texttt{f2(A,B):-move\_up(A,C)}^{216}$
        \item $\texttt{f3(A,B):-move\_up(A,C)}^{1296}$
        \item $\texttt{f4(A,B):-move\_up(A,C)}^{7776}$
    \end{itemize}
    \item By using this approach, \texttt{f0} has to be learnt and added to the BK before \texttt{f1}, \texttt{f1} before \texttt{f2}, \texttt{f2} before \texttt{f3}, \texttt{f3} before \texttt{f4}.
    \item Such, each task will be learnt by searching through hypothesis spaces of sizes as follows:
    \begin{itemize}
        \item \texttt{f0} at size 7
        \item \texttt{f1} at size 8
        \item \texttt{f2} at size 9
        \item \texttt{f3} at size 10
        \item \texttt{f4} at size 11
    \end{itemize}
    \item But, all given tasks will have solutions of 6 body literals:
    \begin{itemize}
        \item $\texttt{f0(A,B):-move\_up(A,C)}^6$
        \item $\texttt{f1(A,B):-f0(A,C)}^{6}$
        \item $\texttt{f2(A,B):-f1(A,C)}^{6}$
        \item $\texttt{f3(A,B):-f2(A,C)}^{6}$
        \item $\texttt{f4(A,B):-f3(A,C)}^{6}$
    \end{itemize}
    \item Given the formula from \autoref{ILPintro} and previous experiments we know that hypothesis spaces containing programs of more than 10 literals can be generally considered unsolvable within a reasonable time limit. Such, we observe the benefit of the new approach.
\end{enumerate}

In the rest of this chapter, we will propose and analyze multiple search strategies and improvements to overcome the aforementioned disadvantages.

\subsection{Knowledge transfer: resetting search size}
We proceed by bringing forward a new search strategy and providing an argument for its theoretical advantages over the current state-of-the-art approach.

When concerned with programs that achieve a significant reduction in size through knowledge transfer, the previous approach performs redundant work by continuing to search for solutions of constantly increasing maximum size. We argue for restarting the search from maximum size to one after a solution is found to one of the tasks.

By doing this we reduce the amount of work required when programs become ``easy" after knowledge transfer is performed. Below we will compute the difference in the number of candidate solutions between the two approaches:
\begin{enumerate}
    \item Let $\mathit{BK}_0$ be our initial background knowledge.
    \item Let $\mathit{BK}_1 = \mathit{BK}_0 \cup \{\texttt{f0}\}$.
    \item Let \texttt{f0}, \texttt{f1} be two predicates such that:
    \begin{itemize}
        \item \texttt{f0} has a solution of size $n$ in $\mathit{BK}_0$.
        \item \texttt{f1} has a solution of size $m > n$ in $\mathit{BK}_0$.
        \item \texttt{f1} has a solution of size $p < m$ in $\mathit{BK}_1$.
    \end{itemize}
    \item Let $\mathit{HS}(\texttt{f}, \mathit{size}, \mathit{BK})$ be the size of the hypothesis space for task \texttt{f}, given $\mathit{size}$ and the background knowledge $\mathit{BK}$.
    \item Let $\mathit{HS}^*(\texttt{f}, \mathit{maxSize}, \mathit{BK}) = \sum\limits_{\mathit{size}=1}^{\mathit{maxSize}}\mathit{HS}(\texttt{f}, \mathit{size}, \mathit{BK})$
    \item Then, the total number of programs generated can be computed as follows:
    \begin{itemize}
        \item Without reset we have: \begin{dmath}\sum\limits_{i=1}^{n}(\mathit{HS}^*(\texttt{f0}, i, \mathit{BK}_0) + \mathit{HS}^*(\texttt{f1}, i, \mathit{BK}_0)) +\\ \mathit{max}(\sum\limits_{i=n + 1}^{p}\mathit{HS}^*(\texttt{f1}, i, \mathit{BK}_1),\mathit{HS}^*(\texttt{f1}, n + 1, \mathit{BK}_1))\end{dmath}
        \item With reset we have: \begin{dmath}\sum\limits_{i=1}^{n}(\mathit{HS}^*(\texttt{f0}, i, \mathit{BK}_0) + \mathit{HS}^*(\texttt{f1}, i, \mathit{BK}_0)) + \sum\limits_{i=1}^{p}\mathit{HS}^*(\texttt{f1}, i, \mathit{BK}_1)
        \end{dmath}
    \end{itemize}
    \item Therefore, we identify two cases:
    \begin{itemize}
        \item If $p \leq n$ (4.1) and (4.2) can be written and compared as follows:
                \begin{dmath}
                    \mathit{HS}^*(\texttt{f1}, n + 1, \mathit{BK}_1) > \sum\limits_{i=1}^{p}\mathit{HS}^*(\texttt{f1}, i, \mathit{BK}_1)
                \end{dmath}
        \item If $p > n$ (4.1) and (4.2) can be written and compared as follows:
            \begin{dmath}
                \sum\limits_{i=n + 1}^{p}\mathit{HS}^*(\texttt{f1}, i, \mathit{BK}_1) < \sum\limits_{i=1}^{p}\mathit{HS}^*(\texttt{f1}, i, \mathit{BK}_1)
            \end{dmath}
    \end{itemize}
    \item By using the formula for $\mathit{HS}$, given in \autoref{ILPintro}, we argue that the performance increase in the case where $p \leq n$ overcomes the performance decrease from the case $p > n$. Moreover, we can extend this argument to multiple tasks (e.g. introduce \texttt{f2} such that \texttt{f2} is very difficult, but has a much simpler solution that uses \texttt{f1}) in order to strengthen the argument.
\end{enumerate}

As a result, we can observe one use case improvement of the reset strategy. 

A more concrete example can be seen in \autoref{fig:reset-example}, where the original approach would have performed redundant work on learning the isGrandgrandmother task by evaluating solutions of size four when a solution can be found at size three. 

\begin{figure}[hbt!]
\begin{minted}{prolog}
    % Background knowledge:
        isFather("John","Dan")
        isFather("Paul","John")

        isWife("Alice","Paul")
        isMother("Olga","Alice")
        
    % Given examples:
        isGrandfather("Paul","Dan")
        
        isGrandmother("Alice","Dan")
        
        isGrandgrandmother("Olga","Dan")
        
    % Programs learnt:
        isGrandfather(A,B) :- isFather(A,C), isFather(C,B)
        
        isGrandmother(A,B) :- isWife(A,C), isGrandfather(C, B)
        
        isGrandgrandmother(A,B) :- isMother(A,C), isGrandmother(C,B)
\end{minted}
\caption{Example for iterative deepening search with search size reset}
\label{fig:reset-example}
\end{figure}

Moreover, we will formalise this claim:
\begin{enumerate}
    \item Assume there exists some task $f$, such that a solution of size $n$ has been found, but there exists a simpler solution, of size $m$, $m < n$. Both solutions generated from the same BK.
    \item By our approach, if a solution of size $n$ has been found then Popper tried all solutions of size $s$, $s\in \{1,2,..,n-1\}$ with the current BK.
    \item Then it means that no solution exists of size $s$, $s \in \{1,2,..,n-1\}$.
    \item Contradiction with $m < n$.
    \item Therefore, $n$ is the smallest solution size that can be found with the current BK.
\end{enumerate}

An \autoref{alg:multi_popper_reset} we showcase pseudo-code for the reset approach.

\begin{algorithm}[hbt!]
\caption{Iterative deepening search with search size reset}\label{alg:multi_popper_reset}
\begin{algorithmic}
\Function{multipopperreset}{}
\State $T \gets \{t_1, t_2,..,t_n\}$
\State $\mathit{BK} \gets \{$BACKGROUND KNOWLEDGE$\}$
\State $S$ $\gets$ \{\}
\State $\mathit{maxSize} \gets 1$
\While{TRUE}
\State $\mathit{change} \gets \mathit{false}$
\For{$t$ in $T$}
    \State $\mathit{sol} \gets$ \Call{popper}{$t$, $\mathit{BK}$, $\mathit{maxSize}$}
    \If{$\mathit{sol} \neq \mathit{null}$}
    \State $\mathit{change}\gets \mathit{true}$
    \State $S$ $\gets$ $S$ $\cup \{\mathit{sol}\}$
    \State $T$ $\gets$ $T$ $\setminus$ \{$t$\} 
    \EndIf
\EndFor
\If{$T =$ \{\}}
\State \Return $S$
\EndIf
\If{$\mathit{change}==\mathit{true}$}
\State $\mathit{maxSize} \gets 1$
\State $\mathit{BK} \gets \mathit{BK} \cup S$
\Else
\State $\mathit{maxSize} \gets \mathit{maxSize} + 1$
\EndIf
\EndWhile
\EndFunction
\end{algorithmic}
\end{algorithm}

\subsection{Knowledge transfer: removing redundant testing}
Recall \autoref{alg:multi_popper_reset}. Any call to Popper would search all solutions up to $\mathit{maxSize}$, each time it is incremented. We can modify the Popper algorithm to only search for programs of a fixed size. Therefore, we will no longer evaluate programs twice unless a solution has been found to one of the tasks. An overview of this change is present in \autoref{alg:multi_popper_reset+}. 
\paragraph{Advantages.} The main advantage of this approach is that it does not reiterate through smaller search space sizes unless necessary (when the BK is augmented).
\paragraph{Disadvantages.} Because we do not re-search the smaller hypothesis space sizes, we do not keep constraint information between searches. This may result in a significant increase in the size of the hypothesis space evaluated. More specifically, if we are searching through a hypothesis space that contains only programs of size $n$, we will not make use of the constraint generated in the previous steps from programs of sizes between $1$ and $n - 1$.
\begin{algorithm}[hbt!]
\caption{Breadth fist search with size reset}\label{alg:multi_popper_reset+}
\begin{algorithmic}
\Function{multipopperreset+}{}
\State $T \gets \{t_1, t_2,..,t_n\}$
\State $\mathit{BK} \gets \{$BACKGROUND KNOWLEDGE$\}$
\State $S$ $\gets$ \{\}
\State $\mathit{size} \gets 1$
\While{TRUE}
\State $\mathit{change} \gets \mathit{false}$
\For{$t$ in $T$}
    \State $\mathit{sol} \gets$ \Call{fixedsizedpopper}{$t$, $\mathit{BK}$, $\mathit{size}$}
    \If{$\mathit{sol} \neq \mathit{null}$}
    \State $\mathit{change}\gets \mathit{true}$
    \State $S$ $\gets$ $S$ $\cup \{\mathit{sol}\}$
    \State $T$ $\gets$ $T$ $\setminus$ \{$t$\} 
    \State \textbf{break}
    \EndIf
\EndFor
\If{$T =$ \{\}}
\State \Return $S$
\EndIf
\If{$\mathit{change}==\mathit{true}$}
\State $\mathit{size} \gets 1$
\State $\mathit{BK} \gets \mathit{BK} \cup S$
\Else
\State $\mathit{size} \gets \mathit{size} + 1$
\EndIf
\EndWhile
\EndFunction
\end{algorithmic}
\end{algorithm}
\section{Preserving constraints}
\label{constraints}
Transferring knowledge between tasks requires Popper to retry hypothesis spaces that contain programs with a smaller number of literals than the maximum tried until now (because the introduction of more programs in the background knowledge creates new possible solutions at the respective sizes). Therefore, all presented approaches will retry the same set of solutions multiple times. 

When modifying search size, the Metagol did not have the ability to transfer knowledge through a standard form, even when working with a single task, e.g. if you have not found a solution of maximum size up to $n$, you would need to restart the search from scratch when looking for a solution of maximum size up to $n + 1$. In this project, we circumvent this limitation by extending the Popper ILP system, which has the ability to retain information from failed attempts in the form of hypothesis constraints.

More specifically, Popper allows us to retain constraints between failed attempts and prune the hypothesis space as soon as we begin searching with different parameters (in our case, an increased maximum number of literals).

To reduce the amount of repeated work caused by search size resets, we adapt all previous methods to preserve constraints learnt by Popper for each task. 

We showcase that necessary changes to implement constraint preservation in the improved reset approach (recall \autoref{alg:multi_popper_reset+}) in \autoref{alg:multi_popper_cons}. 
\begin{algorithm}[hbt!]
\caption{Constraint preserving search}\label{alg:multi_popper_cons}
\begin{algorithmic}
\Function{multipoppercons}{}
\For{$i$ in \Call{range}{n}}
    \State $\mathit{rules}[i] \gets \emptyset$
\EndFor
\State $T \gets \{t_1, t_2,..,t_n\}$
\State $\mathit{BK} \gets \{$BACKGROUND KNOWLEDGE$\}$
\State $S \gets \emptyset$
\State $\mathit{size} \gets 1$
\While{TRUE}
\State $\mathit{change} \gets \mathit{false}$
\For{$t$ in $T$}
    \State $(\mathit{sol},\mathit{cons}) \gets$ \Call{constraintpopper}{$t$, $\mathit{BK}$, $\mathit{size}$, $\mathit{rules}[i]$}
    \State $\mathit{rules}[i] \gets \mathit{rules}[i] \cup \mathit{cons}$
    \If{$\mathit{sol} \neq \mathit{null}$}
    \State $\mathit{change}\gets \mathit{true}$
    \State $S$ $\gets$ $S$ $\cup \{\mathit{sol}\}$
    \State $T$ $\gets$ $T$ $\setminus$ \{$t$\} 
    \State \textbf{break}
    \EndIf
\EndFor
\If{$T =$ \{\}}
\State \Return $S$
\EndIf
\If{$\mathit{change}==\mathit{true}$}
\State $\mathit{size} \gets 1$
\State $\mathit{BK} \gets \mathit{BK} \cup S$
\Else
\State $\mathit{size} \gets \mathit{size} + 1$
\EndIf
\EndWhile
\EndFunction
\end{algorithmic}
\end{algorithm}
\paragraph{Advantages.} By making use of constraint preservation, we remove the bottleneck created by resetting the search on a program and then retesting all candidate hypotheses. Moreover, this technique resolves the significant disadvantage of the breadth first search approach. Now that all constraints are preserved, there is no information lost by not reiterating through the smaller hypothesis spaces.
\paragraph{Disadvantages.} We argue that there do not exist significant disadvantages of preserving constraints. One could point out the fact that they may increase memory requirements and the overhead of the reset phase. However, the expectation is that unless a solution is found early, most of the constraints from previous stages would have also been learnt during the current iteration. 

Finally, we provide an argument for the fact that it is always correct to preserve constraints. 

We are given the fact that any constraint generated by the system is sound \cite{popper}. 

Given a constraint $C$ learnt during an iteration of the learning algorithm with background knowledge $\mathit{BK}$, we show that constraint $C$ would also hold for background knowledge $\mathit{BK}' = \mathit{BK} \cup N$, where $N$ is some set of valid background knowledge. 

Let $h$ be the hypothesis that determined $C$. $h$ is generated from $\mathit{BK}$, therefore  $h$ also generated from $\mathit{BK'}$. Therefore, during a learning process based on $\mathit{BK'}$, if $h$ will also be tested, it will generate the same sound constraint. 
\section{Automatic computation of task learning order}
When considering large data sets where some tasks may be unsolvable, the ordering of the tasks becomes important because the algorithm may not get to consider certain tasks that have a solution of the current search size. Consequently, we propose the introduction of task ordering methods, similar to curriculum learning. We focus on run-time evaluation of tasks, aiming to order tasks by how close our system is to finding a solution. For example, take the following scenario\footnote{For this example we assume that the solution program is always found last in the hypothesis space}:
\begin{enumerate}
    \item We are given 100 tasks, \texttt{fi} with \texttt{i} $\in \{0,..,99\}$.
    \item Task \texttt{f99} has a solution of size seven.
    \item All other tasks have solutions of nine using the initial BK and solutions of size three if \texttt{f99} is part of their BK.
    \item Recall the formula in \autoref{ILPintro}. Based on it, we will approximate the number of programs in hypothesis space of size $x$ to $2^{2^x}$ \footnote{The purpose of this approximation is to provide the reader with an overview of the rate at which the size of the hypothesis space increases with the number of literals. Providing a more accurate result would require a discussion around the values of other parameters from the formula in \autoref{ILPintro}.}. 
    \item Now we will compare the solving time for the reset approach of the best case and worst case orderings.
    \begin{itemize}
        \item Order \texttt{f0}, \texttt{f1},..,\texttt{f99}
        \begin{dmath}
            \sum\limits_{i=1}^{7}(100*2^{2^i})+\sum\limits_{i=1}^{3}(99*2^{2^i})
        \end{dmath}
        \item Order \texttt{f99}, \texttt{f0},\texttt{f1},\texttt{f2},..,\texttt{f98} (\texttt{f99} is computed first)
        \begin{dmath}
            \sum\limits_{i=1}^{6}(100*2^{2^i})+2^{2^7}+\sum\limits_{i=1}^{3}(99*2^{2^i})
        \end{dmath}
    \end{itemize}
    \item We obtain that when the system is given the order \texttt{f0}, \texttt{f1},..,\texttt{f99} it tests approximately 100 times more programs then when it is given the order \texttt{f99}, \texttt{f0},\texttt{f1},\texttt{f2},..,\texttt{f98}.
\end{enumerate}
\paragraph{Advantages.} Given the above example, we know that task ordering is important for the performance of the system. Approaches that perform automatic task ordering can overcome this limitation, by identifying such scenarios and selecting and appropriate task sequence.

\paragraph{Possible disadvantages.} We outline a list of possible disadvantages in order to highlight the points we need to take into account when designing our heuristics. Firstly, the evaluation function is not guaranteed to provide a significant benefit. For example, a heuristic that orders by the number of examples entailed by the best program yet would have no effect on a data set where all tasks are given a single example. Secondly, wrongly estimating the importance of size in the evaluation function may significantly deter search time by testing more extensive programs when not necessary. To mitigate this issue, we experiment with a conservative evaluation function: firstly, ordering by size and then tie-breaking through our heuristic. 

Given the above considerations, we propose two solutions for this problem: one that orders tasks by the number of examples currently solved and the other by the number of constraints generated by the system.
\subsection{Ordering by number of examples solved}
The first approach we experiment with is ordering tasks decreasingly by the number of examples currently solved. We motivate this finding on the fact that very often, entailing an example is equivalent to finding a program that has at least one clause which is part of a solution.
\subsection{Ordering by number of constraints}
The second approach we experiment with is ordering tasks by the number of constraints the system has generated for them. If the system has generated more constraints for a program it means that its hypothesis space can be pruned the most. Therefore, we expect this approach to identify tasks that are closed to being solved.

In \autoref{alg:multi_popper_prio} we provide a generic implementation of the task order approach. It makes use of a priority queue data structure \textit{TPQ}, which is initialised with the desired heuristic.

\begin{algorithm}[hbt!]
\caption{Priority searching of solutions}
\label{alg:multi_popper_prio}
\begin{algorithmic}
\Function{multipopperprio}{}
\For{$i$ in \Call{range}{n}}
    \State $\mathit{rules}[i] \gets \emptyset$
\EndFor
\State $TPQ \gets \{(t_1, 1, 1.0), (t_2, 1, 1.0),..,(t_n, 1, 1.0)\}$
\State $\mathit{BK} \gets \{$BACKGROUND KNOWLEDGE$\}$
\State $S \gets \{\}$
\While{\textbf{not} TPQ \textbf{is empty}}
\State $(t, size, \_)$ $\gets$ \Call{pop}{$TPQ$}
\State $(\mathit{partialsol},\mathit{cons}) \gets$ \Call{popperpartial}{$t$, $BK$, $\mathit{size}$, $\mathit{rules}[i]$}
\State $\mathit{rules}[i] \gets \mathit{rules}[i] \cup \mathit{cons}$
\If{\Call{isvalidsol}{$\mathit{partialsol}$,$t$}}
\State $S$ $\gets$ $S$ $\cup \{\mathit{sol}\}$
\State $\mathit{BK} \gets \mathit{BK} \cup S$
\State \Call{ResetPopper}{T}
\State $T$ $\gets$ $T$ $\setminus$ \{$t$\} 
\State $TPQ \gets \{(t_i, 1, 1.0) \vert t_i \in T \}$
\Else
\If{$partialsol \neq null$}
\State $priority \gets$ \Call{computeprio}{$sol$}
\State \Call{PUSH}{$TPQ$, $(t, size + 1, priority)$}
\EndIf
\EndIf
\EndWhile
\State \Return $S$
\EndFunction
\end{algorithmic}
\end{algorithm}
\chapter{Experimental Results}
The goal of this chapter is to determine the performance of the approaches proposed in the previous chapter. We will perform experiments in order to obtain answers to the following questions: 

\textbf{Q1.} Does constraint preservation offer an improvement in learning speed for approaches that repeat searching over the same hypothesis space? 

For each strategy mentioned previously, we will evaluate the running time change resulting from enabling constraint preservation on multiple data sets that present a wide array of specific features. 

\textbf{Q2.} Do the advantages of resetting search size overcome its disadvantages when compared to the state-of-the-art approach? 

We will compare the two reset strategies with the current state-of-the-art iterative deepening search approach on data sets that are expected to benefit each method and compare the best case and worst case results for each algorithm. 

\textbf{Q3.} Is there any benefit in heuristically ordering programs on the same size level during a search? 

We will compare the two ordering strategies against a system that selects a random ordering at the beginning of the learning process and respects it throughout each stage. We have chosen this benchmark in order to remove any bias from a system that takes the given original order, which may be beneficial/not beneficial. 

The scope of this experiment will be to determine if heuristic program ordering is either beneficial, neutral, or provides a drawback through the added overhead.
\section{Data sets}
\textbf{Strings transformation.} We test our system using an existing data set of string transformation operations, out of which we select the ones that have been found solvable by Popper during previous experiments \cite{string-data}.

\textbf{Robot movement.} We generate a set of tasks that focus on robot movement. The goal of each task is to learn to traverse a large area using only unit length movement instructions. This task will be used to compare the performance of the reset strategy and state-of-the-art strategy on a large set of tasks that can all be reduced to a size of three literals in a multi-task learning setting. 

\textbf{Printer head.} We generate a set of tasks that simulate the movement of a printer head over a piece of paper. The printer head is able to perform one pixel size movements, fill a pixel with colour or return to an initial position. 

This data set is an extension of the robot movement set, which also allows recursion. Instead of learning to traverse fixed distances, the printer head learns to work on sheets of arbitrary size. 

\begin{figure}[!htb]
    \centering
    \begin{minipage}{.5\textwidth}
        \centering
        \includegraphics[width=0.4\linewidth]{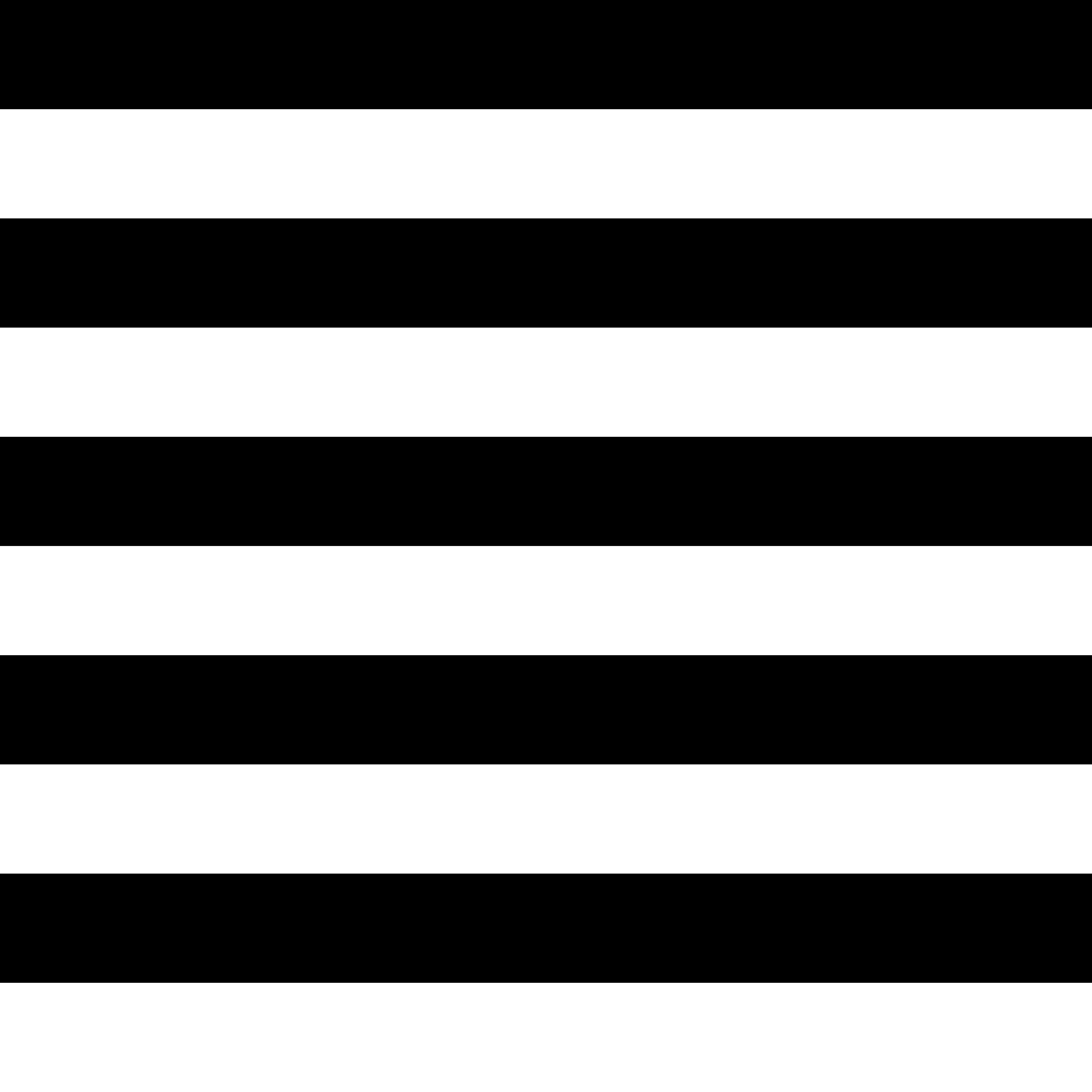}
        \caption{Zebra pattern image}
        \label{fig:zebra}
    \end{minipage}%
    \begin{minipage}{0.5\textwidth}
        \centering
        \includegraphics[width=0.4\linewidth]{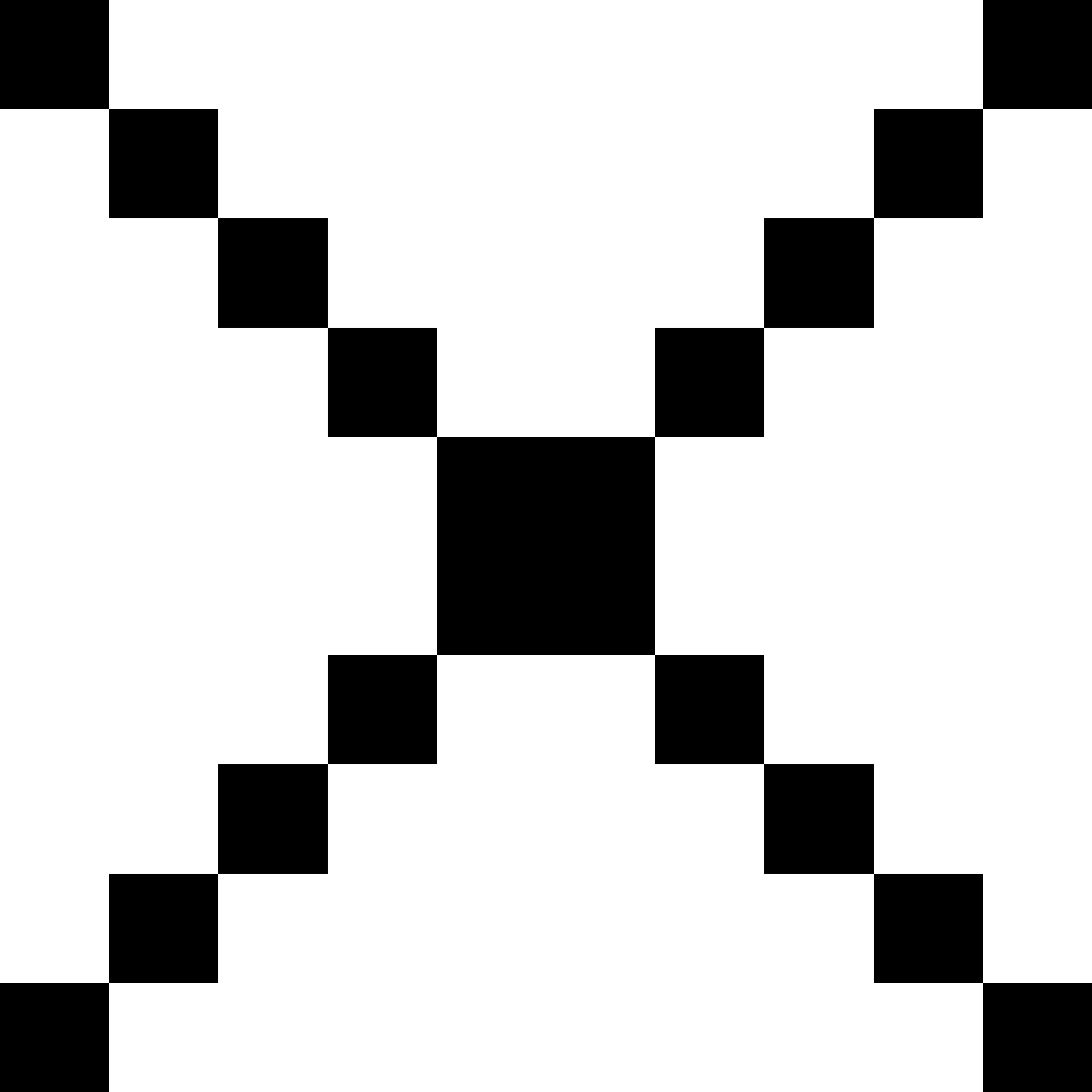}
    \caption{Cross pattern image}
    \label{fig:cross}
    \end{minipage}
\end{figure}

In \autoref{fig:zebra} and \autoref{fig:cross} we show two examples of pixel images used in training the system.
\section{Testing environment}
\subsection{Specifications}
All experiments have been run on Amazon EC2 machines (c6i.large), which have the following specifications: 

\textbf{Operating system:} Ubuntu 20.04.3 LTS

\textbf{Software versions:}
\begin{itemize}
    \item \textbf{Python} 3.8.10
    \item \textbf{PySwip} 0.2.11
    \item \textbf{Clingo} 5.5.0
    \item \textbf{SWI-Prolog} 8.4.2
\end{itemize}
\textbf{Processor:} Intel(R) Xeon(R) Platinum 8375C CPU @ 2.90GHz 

\textbf{Number of cores:} 2 

\textbf{Threads per cores:} 2 

\textbf{RAM memory:} 4 GB
\subsection{Testing protocol}
For all data sets, each strategy was run ten times with a timeout limit of 6000 seconds. 

Before each run, the ordering of the input tasks is randomly determined. This approach allows our results to ensure the reliability of our data and provide greater confidence that the observed performance of the system is consistent. 

There are significant sources of noise that require experiments to be run multiple times. For example, some input orderings benefit some systems over others. Furthermore, the Popper system searches through a hypothesis space in a non-deterministic manner. Therefore, during different runs, specific tasks can be given different definitions that are not necessarily reusable in the same contexts.
\section{Results}
\subsection{Algorithms}
In this section, we will define a list of acronyms for each of our search strategies in order to make it easier for the reader to interpret the results. 

\textbf{NAIVE} is the naive algorithm. 

\textbf{ID} is the state-of-the-art, iterative deepening approach. 

\textbf{RESET-ID} is the iterative deepening reset approach. 

\textbf{RESET-BFS} is the breadth first search reset approach. 

\textbf{PRIO-EX} is a version of RESET-BFS that sorts tasks on the same level decreasingly by the number of examples solved. 

\textbf{PRIO-CONS} is a version of RESET-BFS that sorts tasks on the same level decreasingly by the number of constraints discovered. 

If any algorithm name is followed by the PRESERVE keyword (e.g. ID PRESERVE), it means that constraint preservation option has been set true. Otherwise, we assume that constraint preservation has not been used. 

Because the breadth first search based algorithms perform poorly without constraint preservation, we will choose always to enable constraint preservation when comparing the performance of multiple strategies.
\subsection{Printer head}
\subsubsection{Overview}
In \autoref{fig:printer-results} we showcase the experimental results for all our algorithms as a percentage of tasks solved after several minutes. This data set contains many tasks that are solved relatively early and a few complicated tasks, such we attach an additional plot (see \autoref{fig:printer-sec-results}) that highlights the behaviour of the strategies in the first 100 seconds. We consider that this metric better describes the performance of the systems because it more clearly showcases their differences and removes the redundant noise generated by task ordering in the higher hypothesis spaces.
\pgfplotsset{
    myplotstyle/.style={
    legend style={draw=none, font=\small},
    legend cell align=left,
    legend pos=outer north east,
    ylabel style={align=center, font=\bfseries\boldmath},
    xlabel style={align=center, font=\bfseries\boldmath},
    x tick label style={font=\bfseries\boldmath},
    y tick label style={font=\bfseries\boldmath},
    ymax = 100,
    scaled ticks=false,
    every axis plot/.append style={thick},
    },
}
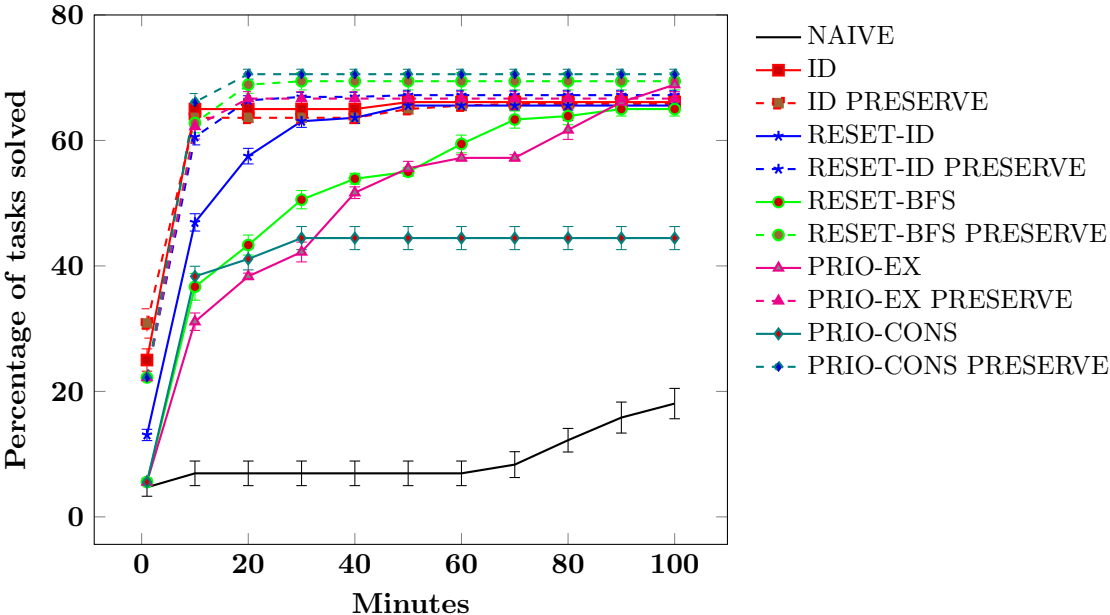
\begin{figure}[hbt!]
\centering
\begin{tikzpicture}
\begin{axis}[
    myplotstyle,
    legend pos=outer north east,
    legend entries={NAIVE, ID, ID PRESERVE, RESET-ID, RESET-ID PRESERVE, RESET-BFS, RESET-BFS PRESERVE,
    PRIO-EX, PRIO-EX PRESERVE, PRIO-CONS, PRIO-CONS PRESERVE},
    xlabel={Minutes},
    ylabel={Percentage of tasks solved},
    ymax = 80,
    width=10cm
]
\addplot+[mark = none, color = black, solid,error bars/.cd, y dir=both,y explicit] table[x=Minutes,y=NAIVE, col sep=comma, y error = NAIVE ERROR]{anc/Results-Images_Min.csv};
\addplot+[mark = square*, color = red, solid,error bars/.cd, y dir=both,y explicit] table[x=Minutes,y=SOTA NO-PRESERVE, col sep=comma, y error = SOTA NO-PRESERVE ERROR]{anc/Results-Images_Min.csv};
\addplot+[mark = square*, color = red, dashed,error bars/.cd, y dir=both,y explicit] table[x=Minutes,y=SOTA PRESERVE, col sep=comma,y error = SOTA PRESERVE ERROR]{anc/Results-Images_Min.csv};
\addplot+[mark = star, color = blue, solid,error bars/.cd, y dir=both,y explicit] table[x=Minutes,y=RESET NO-PRESERVE, col sep=comma, y error = RESET NO-PRESERVE ERROR]{anc/Results-Images_Min.csv};
\addplot+[mark = star, color = blue, dashed,error bars/.cd, y dir=both,y explicit] table[x=Minutes,y=RESET PRESERVE, col sep=comma, y error = RESET PRESERVE ERROR]{anc/Results-Images_Min.csv};
\addplot+[mark = *, color = green, solid,error bars/.cd, y dir=both,y explicit] table[x=Minutes,y=RESET-IMP NO-PRESERVE, col sep=comma, y error = RESET-IMP NO-PRESERVE ERROR]{anc/Results-Images_Min.csv};
\addplot+[mark = *, color = green, dashed,error bars/.cd, y dir=both,y explicit] table[x=Minutes,y=RESET-IMP PRESERVE, col sep=comma, y error = RESET-IMP PRESERVE ERROR]{anc/Results-Images_Min.csv};
\addplot+[mark = triangle*, color = magenta, solid,error bars/.cd, y dir=both,y explicit] table[x=Minutes,y=PRIO NO-PRESERVE, col sep=comma, y error = PRIO NO-PRESERVE ERROR]{anc/Results-Images_Min.csv};
\addplot+[mark = triangle*, color = magenta, dashed,error bars/.cd, y dir=both,y explicit] table[x=Minutes,y=PRIO PRESERVE, col sep=comma, y error = PRIO PRESERVE ERROR]{anc/Results-Images_Min.csv};
\addplot+[mark = diamond*, color = teal, solid,error bars/.cd, y dir=both,y explicit] table[x=Minutes,y=PRIO CONS NO-PRESERVE, col sep=comma, y error = PRIO CONS NO-PRESERVE ERROR]{anc/Results-Images_Min.csv};
\addplot+[mark = diamond*, color = teal, dashed,error bars/.cd, y dir=both,y explicit] table[x=Minutes,y=PRIO CONS PRESERVE, col sep=comma, y error = PRIO CONS PRESERVE ERROR]{anc/Results-Images_Min.csv};
\end{axis}
\end{tikzpicture}
    \caption{Overview of printer head results (showing standard error)}
    \label{fig:printer-results}
\end{figure}

\begin{figure}[hbt!]
\centering
\begin{tikzpicture}
\begin{axis}[
    myplotstyle,
    legend pos=outer north east,
    legend entries={NAIVE, ID, ID PRESERVE, RESET-ID, RESET-ID PRESERVE, RESET-BFS, RESET-BFS PRESERVE,
    PRIO-EX, PRIO-EX PRESERVE, PRIO-CONS, PRIO-CONS PRESERVE},
    xlabel={Seconds},
    ylabel={Percentage of tasks solved},
    ymax = 60,
    width=10cm
]
\addplot+[mark = none, color = black, solid,error bars/.cd, y dir=both,y explicit] table[x=Minutes,y=NAIVE, col sep=comma, y error = NAIVE ERROR]{anc/Results-Images_Sec.csv};
\addplot+[mark = square*, color = red, solid,error bars/.cd, y dir=both,y explicit] table[x=Minutes,y=SOTA NO-PRESERVE, col sep=comma, y error = SOTA NO-PRESERVE ERROR]{anc/Results-Images_Sec.csv};
\addplot+[mark = square*, color = red, dashed,error bars/.cd, y dir=both,y explicit] table[x=Minutes,y=SOTA PRESERVE, col sep=comma,y error = SOTA PRESERVE ERROR]{anc/Results-Images_Sec.csv};
\addplot+[mark = star, color = blue, solid,error bars/.cd, y dir=both,y explicit] table[x=Minutes,y=RESET NO-PRESERVE, col sep=comma, y error = RESET NO-PRESERVE ERROR]{anc/Results-Images_Sec.csv};
\addplot+[mark = star, color = blue, dashed,error bars/.cd, y dir=both,y explicit] table[x=Minutes,y=RESET PRESERVE, col sep=comma, y error = RESET PRESERVE ERROR]{anc/Results-Images_Sec.csv};
\addplot+[mark = *, color = green, solid,error bars/.cd, y dir=both,y explicit] table[x=Minutes,y=RESET-IMP NO-PRESERVE, col sep=comma, y error = RESET-IMP NO-PRESERVE ERROR]{anc/Results-Images_Sec.csv};
\addplot+[mark = *, color = green, dashed,error bars/.cd, y dir=both,y explicit] table[x=Minutes,y=RESET-IMP PRESERVE, col sep=comma, y error = RESET-IMP PRESERVE ERROR]{anc/Results-Images_Sec.csv};
\addplot+[mark = triangle*, color = magenta, solid,error bars/.cd, y dir=both,y explicit] table[x=Minutes,y=PRIO NO-PRESERVE, col sep=comma, y error = PRIO NO-PRESERVE ERROR]{anc/Results-Images_Sec.csv};
\addplot+[mark = triangle*, color = magenta, dashed,error bars/.cd, y dir=both,y explicit] table[x=Minutes,y=PRIO PRESERVE, col sep=comma, y error = PRIO PRESERVE ERROR]{anc/Results-Images_Sec.csv};
\addplot+[mark = diamond*, color = teal, solid,error bars/.cd, y dir=both,y explicit] table[x=Minutes,y=PRIO CONS NO-PRESERVE, col sep=comma, y error = PRIO CONS NO-PRESERVE ERROR]{anc/Results-Images_Sec.csv};
\addplot+[mark = diamond*, color = teal, dashed,error bars/.cd, y dir=both,y explicit] table[x=Minutes,y=PRIO CONS PRESERVE, col sep=comma, y error = PRIO CONS PRESERVE ERROR]{anc/Results-Images_Sec.csv};
\end{axis}
\end{tikzpicture}
    \caption{Zoom in of printer head results for the first 100 seconds (showing standard error)}
    \label{fig:printer-sec-results}
\end{figure}

 \begin{table}[hbt!]
 \centering
\begin{tabular}{c|c|c|c}
        Algorithm & \multicolumn{1}{l|}{Min} & \multicolumn{1}{l|}{Max} & \multicolumn{1}{l}{Standard deviation} \\ \hline
NAIVE              & 1                        & 9                       & 1.97                                   \\
PRIO-CONS          & 7                       & 10                       & 1.1                                   \\
ID               & 11                       & 13                       & 0.7                                   \\
ID PRESERVE      & 11                       & 13                       & 0.65                                    \\
RESET-ID           & 11                       & 13                       & 0.6                                   \\
RESET-ID PRESERVE  & 11                        & 13                       & 0.62                                  \\
RESET-BFS PRESERVE & 11                       & 13                       & 0.64                                  \\
PRIO-EX            & 11                       & 13                       & 0.66                                   \\
PRIO-EX PRESERVE   & 11                       & 13                       & 0.63                                    \\
PRIO-CONS PRESERVE & 12                       & 13                       & 0.46                                  \\
RESET-BFS          & 11                       & 14                       & 0.81                                  
\end{tabular}
    \caption{Printer head algorithms results variance}
    \label{tab:printer-table}
\end{table}

\subsubsection{Question 1}
\begin{figure}[hbt!]
\centering
\begin{tikzpicture}
\begin{axis}[
    myplotstyle,
    legend pos=outer north east,
    legend entries={ID, ID PRESERVE, RESET-ID, RESET-ID PRESERVE, RESET-BFS, RESET-BFS PRESERVE},
    xlabel={Seconds},
    ylabel={Percentage of tasks solved},
    ymax = 60,
    width=10cm
]
\addplot+[mark = square*, color = red, solid,error bars/.cd, y dir=both,y explicit] table[x=Minutes,y=SOTA NO-PRESERVE, col sep=comma, y error = SOTA NO-PRESERVE ERROR]{anc/Results-Images_Sec.csv};
\addplot+[mark = square*, color = red, dashed,error bars/.cd, y dir=both,y explicit] table[x=Minutes,y=SOTA PRESERVE, col sep=comma,y error = SOTA PRESERVE ERROR]{anc/Results-Images_Sec.csv};
\addplot+[mark = star, color = blue, solid,error bars/.cd, y dir=both,y explicit] table[x=Minutes,y=RESET NO-PRESERVE, col sep=comma, y error = RESET NO-PRESERVE ERROR]{anc/Results-Images_Sec.csv};
\addplot+[mark = star, color = blue, dashed,error bars/.cd, y dir=both,y explicit] table[x=Minutes,y=RESET PRESERVE, col sep=comma, y error = RESET PRESERVE ERROR]{anc/Results-Images_Sec.csv};
\addplot+[mark = *, color = green, solid,error bars/.cd, y dir=both,y explicit] table[x=Minutes,y=RESET-IMP NO-PRESERVE, col sep=comma, y error = RESET-IMP NO-PRESERVE ERROR]{anc/Results-Images_Sec.csv};
\addplot+[mark = *, color = green, dashed,error bars/.cd, y dir=both,y explicit] table[x=Minutes,y=RESET-IMP PRESERVE, col sep=comma, y error = RESET-IMP PRESERVE ERROR]{anc/Results-Images_Sec.csv};
\end{axis}
\end{tikzpicture}
    \caption{Preserve comparison of printer head results for the first 100 seconds (showing standard error)}
    \label{fig:q1-printer-results}
\end{figure}
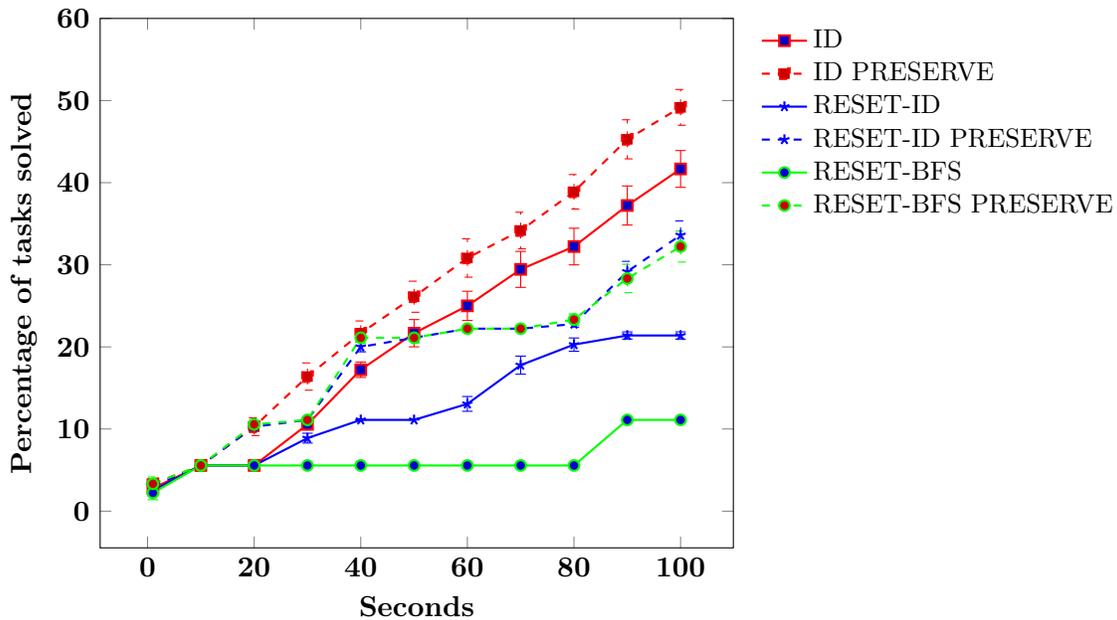

In \autoref{fig:printer-results} and \autoref{fig:q1-printer-results} we observe that by enabling the preserve option, all algorithms consistently outperform their default variants. Therefore, we conclude that the answer for \textbf{Q1} is \textbf{yes}. This result confirms our expectation that constraint preservation will result in at least a minor improvement in all use cases.
\subsubsection{Question 2}
\begin{figure}[hbt!]
\centering
\begin{tikzpicture}
\begin{axis}[
    myplotstyle,
    legend pos=outer north east,
    legend entries={ID PRESERVE, RESET-ID PRESERVE, RESET-BFS PRESERVE},
    xlabel={Seconds},
    ylabel={Percentage of tasks solved},
    ymax = 60,
    width=10cm
]
\addplot+[mark = square*, color = red, dashed,error bars/.cd, y dir=both,y explicit] table[x=Minutes,y=SOTA PRESERVE, col sep=comma,y error = SOTA PRESERVE ERROR]{anc/Results-Images_Sec.csv};
\addplot+[mark = star, color = blue, dashed,error bars/.cd, y dir=both,y explicit] table[x=Minutes,y=RESET PRESERVE, col sep=comma, y error = RESET PRESERVE ERROR]{anc/Results-Images_Sec.csv};
\addplot+[mark = *, color = green, dashed,error bars/.cd, y dir=both,y explicit] table[x=Minutes,y=RESET-IMP PRESERVE, col sep=comma, y error = RESET-IMP PRESERVE ERROR]{anc/Results-Images_Sec.csv};
\end{axis}
\end{tikzpicture}
    \caption{Graph comparing state-of-the-art and reset strategies for printer data during the first 100 seconds (showing standard error)}
    \label{fig:q2-printer-results}
\end{figure}
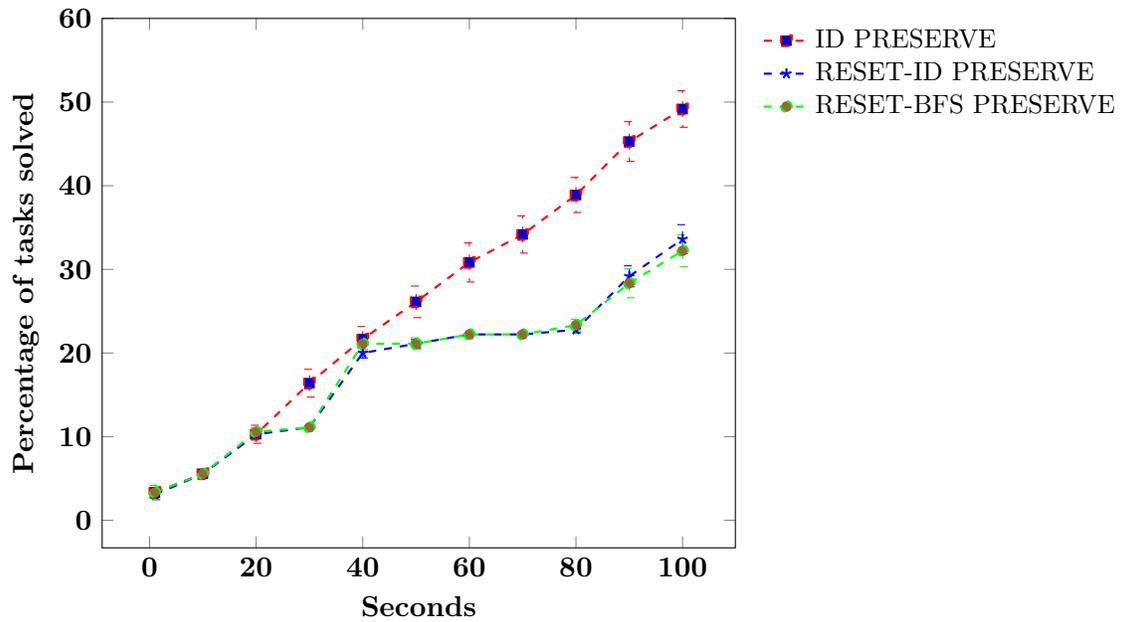

By analysing \autoref{fig:q2-printer-results} and \autoref{fig:printer-results} we observe that the reset approaches perform slightly worse when they have to perform reset operations quite often (solving simple tasks repeatedly), but end up performing similarly to the state-of-the-art approach in the long run. Therefore, we determine that the answer for \textbf{Q2} is \textbf{no}.
\subsubsection{Question 3}
We do not observe any significant benefit of the priority approaches that can not be attributed to noise. Therefore, we conclude that the answer for \textbf{Q3} is \textbf{no}. It is worth mentioning that both priority algorithms compute the most difficult task faster on average than the other approaches. Their average times to learn 13 tasks are 661 and 671 seconds, whereas the state-of-the-art approach achieves a mean of 1951 seconds and the best reset approach 1096. Is it highly likely that these results are a consequence of variation in the initial ordering and do not reflect the actual performance of the search strategies.
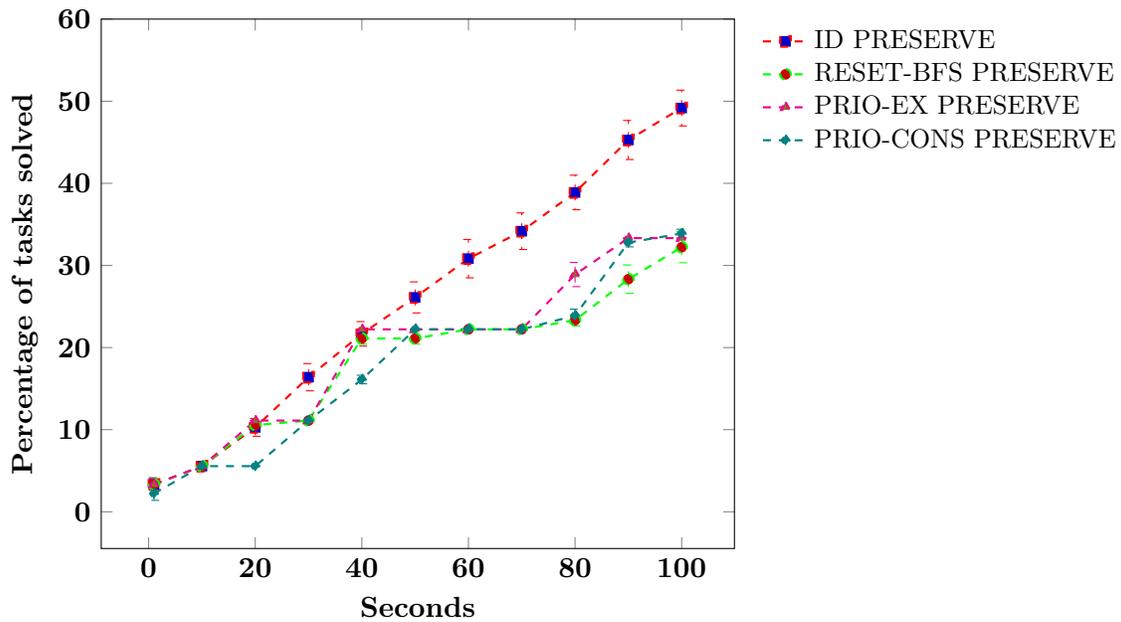
\begin{figure}[hbt!]
\centering
\begin{tikzpicture}
\begin{axis}[
    myplotstyle,
    legend pos=outer north east,
    legend entries={ID PRESERVE, RESET-BFS PRESERVE,
    PRIO-EX PRESERVE, PRIO-CONS PRESERVE},
    xlabel={Seconds},
    ylabel={Percentage of tasks solved},
    ymax = 60,
    width=10cm
]
\addplot+[mark = square*, color = red, dashed,error bars/.cd, y dir=both,y explicit] table[x=Minutes,y=SOTA PRESERVE, col sep=comma,y error = SOTA PRESERVE ERROR]{anc/Results-Images_Sec.csv};
\addplot+[mark = *, color = green, dashed,error bars/.cd, y dir=both,y explicit] table[x=Minutes,y=RESET-IMP PRESERVE, col sep=comma, y error = RESET-IMP PRESERVE ERROR]{anc/Results-Images_Sec.csv};
\addplot+[mark = triangle*, color = magenta, dashed,error bars/.cd, y dir=both,y explicit] table[x=Minutes,y=PRIO PRESERVE, col sep=comma, y error = PRIO PRESERVE ERROR]{anc/Results-Images_Sec.csv};
\addplot+[mark = diamond*, color = teal, dashed,error bars/.cd, y dir=both,y explicit] table[x=Minutes,y=PRIO CONS PRESERVE, col sep=comma, y error = PRIO CONS PRESERVE ERROR]{anc/Results-Images_Sec.csv};
\end{axis}
\end{tikzpicture}
    \caption{Zoom in of printer head results for the first 100 seconds (showing standard error)}
    \label{fig:q3-printer-results}
\end{figure}
\subsection{String transformation}
\subsubsection{Overview}
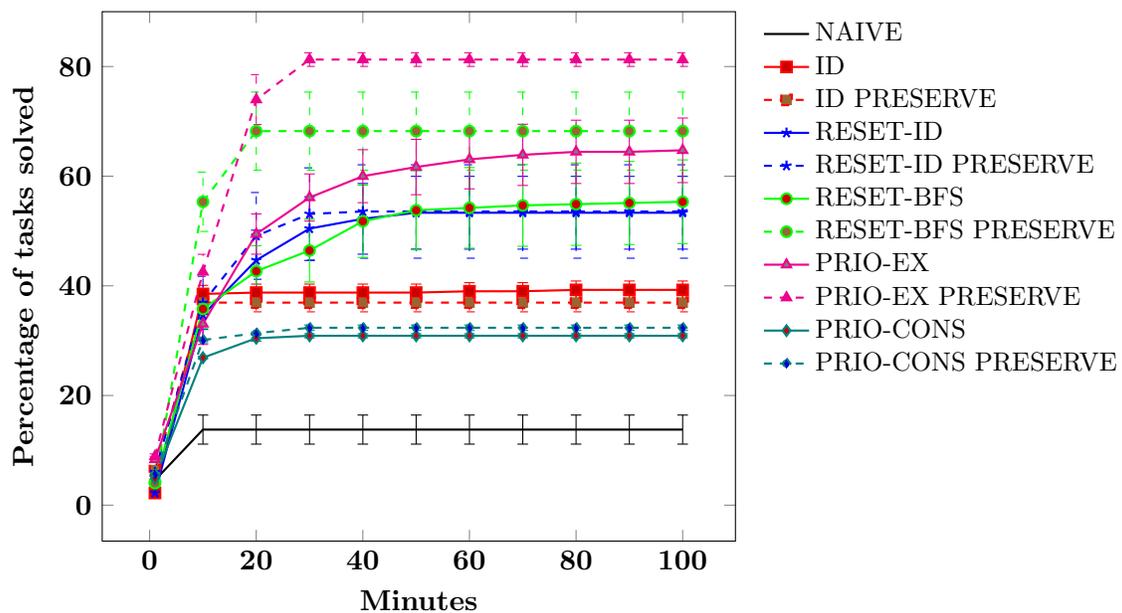
\begin{figure}[hbt!]
\centering
\begin{tikzpicture}
\begin{axis}[
    myplotstyle,
    legend pos=outer north east,
    legend entries={NAIVE, ID, ID PRESERVE, RESET-ID, RESET-ID PRESERVE, RESET-BFS, RESET-BFS PRESERVE,
    PRIO-EX, PRIO-EX PRESERVE, PRIO-CONS, PRIO-CONS PRESERVE},
    xlabel={Minutes},
    ylabel={Percentage of tasks solved},
    ymax = 90,
    width=10cm
]
\addplot+[mark = none, color = black, solid,error bars/.cd, y dir=both,y explicit] table[x=Minutes,y=NAIVE, col sep=comma, y error = NAIVE ERROR]{anc/Results-Strings_Min.csv};
\addplot+[mark = square*, color = red, solid,error bars/.cd, y dir=both,y explicit] table[x=Minutes,y=SOTA NO-PRESERVE, col sep=comma, y error = SOTA NO-PRESERVE ERROR]{anc/Results-Strings_Min.csv};
\addplot+[mark = square*, color = red, dashed,error bars/.cd, y dir=both,y explicit] table[x=Minutes,y=SOTA PRESERVE, col sep=comma,y error = SOTA PRESERVE ERROR]{anc/Results-Strings_Min.csv};
\addplot+[mark = star, color = blue, solid,error bars/.cd, y dir=both,y explicit] table[x=Minutes,y=RESET NO-PRESERVE, col sep=comma, y error = RESET NO-PRESERVE ERROR]{anc/Results-Strings_Min.csv};
\addplot+[mark = star, color = blue, dashed,error bars/.cd, y dir=both,y explicit] table[x=Minutes,y=RESET PRESERVE, col sep=comma, y error = RESET PRESERVE ERROR]{anc/Results-Strings_Min.csv};
\addplot+[mark = *, color = green, solid,error bars/.cd, y dir=both,y explicit] table[x=Minutes,y=RESET-IMP NO-PRESERVE, col sep=comma, y error = RESET-IMP NO-PRESERVE ERROR]{anc/Results-Strings_Min.csv};
\addplot+[mark = *, color = green, dashed,error bars/.cd, y dir=both,y explicit] table[x=Minutes,y=RESET-IMP PRESERVE, col sep=comma, y error = RESET-IMP PRESERVE ERROR]{anc/Results-Strings_Min.csv};
\addplot+[mark = triangle*, color = magenta, solid,error bars/.cd, y dir=both,y explicit] table[x=Minutes,y=PRIO NO-PRESERVE, col sep=comma, y error = PRIO NO-PRESERVE ERROR]{anc/Results-Strings_Min.csv};
\addplot+[mark = triangle*, color = magenta, dashed,error bars/.cd, y dir=both,y explicit] table[x=Minutes,y=PRIO PRESERVE, col sep=comma, y error = PRIO PRESERVE ERROR]{anc/Results-Strings_Min.csv};
\addplot+[mark = diamond*, color = teal, solid,error bars/.cd, y dir=both,y explicit] table[x=Minutes,y=PRIO CONS NO-PRESERVE, col sep=comma, y error = PRIO CONS NO-PRESERVE ERROR]{anc/Results-Strings_Min.csv};
\addplot+[mark = diamond*, color = teal, dashed,error bars/.cd, y dir=both,y explicit] table[x=Minutes,y=PRIO CONS PRESERVE, col sep=comma, y error = PRIO CONS PRESERVE ERROR]{anc/Results-Strings_Min.csv};
\end{axis}
\end{tikzpicture}
    \caption{Overview of string transformation results (showing standard error)}
    \label{fig:string-results}
\end{figure}

In \autoref{fig:string-results} we present an overview of the performance of each algorithm, expressed as the percentage of tasks solved after several minutes. We also outline the variance in \autoref{tab:string-table}, by listing their worst and best performances, along with the standard deviation of the total number of tasks solved.

We observe the fact that no algorithm was able to solve all the tasks. We attribute this to the fact that we have used functional testing in order to validate if a generated program is correct, whereas the original experiment did not enable this option. Functional testing refers to evaluating predicates by disregarding their truth value and only considering their output \cite{functional-definition}.

 \begin{table}[hbt!]
 \centering
\begin{tabular}{c|c|c|c}
        Algorithm & \multicolumn{1}{l|}{Min} & \multicolumn{1}{l|}{Max} & \multicolumn{1}{l}{Standard deviation} \\ \hline
PRIO-CONS          & 13                       & 15                       & 0.54                                   \\
PRIO-CONS PRESERVE & 14                       & 16                       & 0.83                                    \\
ID PRESERVE      & 12                       & 19                       & 2.5                                    \\
NAIVE              & 0                        & 21                       & 6.28                                   \\
ID               & 12                       & 21                       & 2.45                                   \\
RESET-BFS          & 11                       & 38                       & 11.41                                  \\
RESET-ID           & 13                       & 39                       & 9.78                                   \\
PRIO-EX            & 20                       & 39                       & 7.87                                   \\
RESET-BFS PRESERVE & 15                       & 40                       & 10.39                                  \\
PRIO-EX PRESERVE   & 35                       & 40                       & 1.5                                    
\end{tabular}
    \caption{String transformation algorithms results variance
    }
    \label{tab:string-table}
\end{table}
\subsubsection{Question 1}
\begin{figure}[hbt!]
\centering
\begin{tikzpicture}
\begin{axis}[
    myplotstyle,
    legend pos=outer north east,
    legend entries={ID, ID PRESERVE, RESET-ID, RESET-ID PRESERVE, RESET-BFS, RESET-BFS PRESERVE},
    xlabel={Minutes},
    ylabel={Percentage of tasks solved},
    ymax = 90,
    width=10cm
]
\addplot+[mark = square*, color = red, solid,error bars/.cd, y dir=both,y explicit] table[x=Minutes,y=SOTA NO-PRESERVE, col sep=comma, y error = SOTA NO-PRESERVE ERROR]{anc/Results-Strings_Min.csv};
\addplot+[mark = square*, color = red, dashed,error bars/.cd, y dir=both,y explicit] table[x=Minutes,y=SOTA PRESERVE, col sep=comma,y error = SOTA PRESERVE ERROR]{anc/Results-Strings_Min.csv};
\addplot+[mark = star, color = blue, solid,error bars/.cd, y dir=both,y explicit] table[x=Minutes,y=RESET NO-PRESERVE, col sep=comma, y error = RESET NO-PRESERVE ERROR]{anc/Results-Strings_Min.csv};
\addplot+[mark = star, color = blue, dashed,error bars/.cd, y dir=both,y explicit] table[x=Minutes,y=RESET PRESERVE, col sep=comma, y error = RESET PRESERVE ERROR]{anc/Results-Strings_Min.csv};
\addplot+[mark = *, color = green, solid,error bars/.cd, y dir=both,y explicit] table[x=Minutes,y=RESET-IMP NO-PRESERVE, col sep=comma, y error = RESET-IMP NO-PRESERVE ERROR]{anc/Results-Strings_Min.csv};
\addplot+[mark = *, color = green, dashed,error bars/.cd, y dir=both,y explicit] table[x=Minutes,y=RESET-IMP PRESERVE, col sep=comma, y error = RESET-IMP PRESERVE ERROR]{anc/Results-Strings_Min.csv};
\end{axis}
\end{tikzpicture}
    \caption{Graph showing the effects of the preserve option for string data (showing standard error)}
    \label{fig:q1-string-results}
\end{figure}
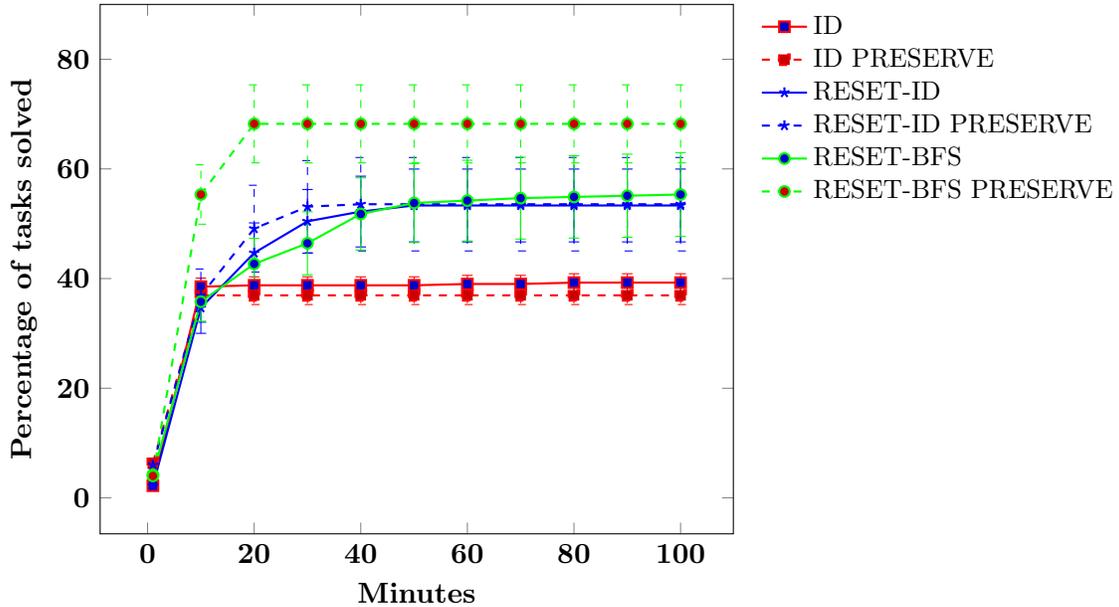

For the first question, we compare the results of the three main search strategies with the constraint preservation option both enabled and disabled. The result of the comparison is present in \autoref{fig:q1-string-results}. We conclude that the answer to \textbf{Q1} is \textbf{yes}, as all strategies perform at least as well as their default version when constraint preservation is used.
\subsubsection{Question 2}
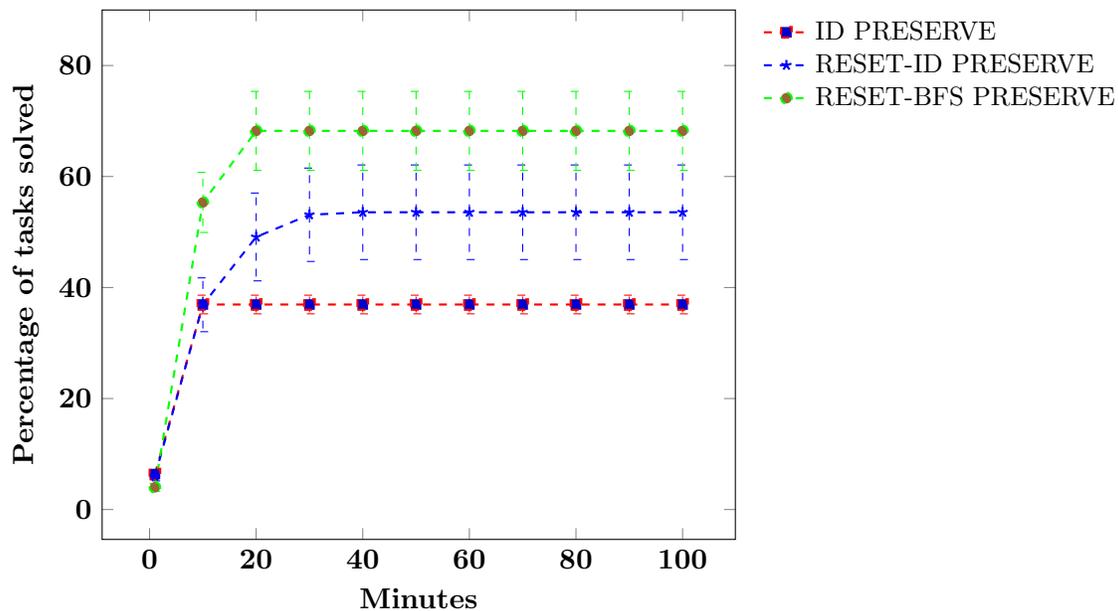
\begin{figure}[hbt!]
\centering
\begin{tikzpicture}
\begin{axis}[
    myplotstyle,
    legend pos=outer north east,
    legend entries={ID PRESERVE, RESET-ID PRESERVE, RESET-BFS PRESERVE},
    xlabel={Minutes},
    ylabel={Percentage of tasks solved},
    ymax = 90,
    width=10cm
]
\addplot+[mark = square*, color = red, dashed,error bars/.cd, y dir=both,y explicit] table[x=Minutes,y=SOTA PRESERVE, col sep=comma,y error = SOTA PRESERVE ERROR]{anc/Results-Strings_Min.csv};
\addplot+[mark = star, color = blue, dashed,error bars/.cd, y dir=both,y explicit] table[x=Minutes,y=RESET PRESERVE, col sep=comma, y error = RESET PRESERVE ERROR]{anc/Results-Strings_Min.csv};
\addplot+[mark = *, color = green, dashed,error bars/.cd, y dir=both,y explicit] table[x=Minutes,y=RESET-IMP PRESERVE, col sep=comma, y error = RESET-IMP PRESERVE ERROR]{anc/Results-Strings_Min.csv};
\end{axis}
\end{tikzpicture}
    \caption{Graph comparing state-of-the-art and reset strategies for string data (showing standard error)}
    \label{fig:q2-string-results}
\end{figure}

In \autoref{fig:q2-string-results} we compare the results of the state-of-the-art strategy, versus the two reset strategies we suggested. Since both reset strategies perform better than the state-of-the-art approach we conclude that the answer for \textbf{Q2} is \textbf{yes}.
\subsubsection{Question 3}
\begin{figure}[hbt!]
\centering
\begin{tikzpicture}
\begin{axis}[
    myplotstyle,
    legend pos=outer north east,
    legend entries={ID PRESERVE, RESET-BFS PRESERVE, PRIO-EX PRESERVE, PRIO-CONS PRESERVE},
    xlabel={Minutes},
    ylabel={Percentage of tasks solved},
    ymax = 90,
    width=10cm
]
\addplot+[mark = square*, color = red, dashed,error bars/.cd, y dir=both,y explicit] table[x=Minutes,y=SOTA PRESERVE, col sep=comma,y error = SOTA PRESERVE ERROR]{anc/Results-Strings_Min.csv};
\addplot+[mark = *, color = green, dashed,error bars/.cd, y dir=both,y explicit] table[x=Minutes,y=RESET-IMP PRESERVE, col sep=comma, y error = RESET-IMP PRESERVE ERROR]{anc/Results-Strings_Min.csv};
\addplot+[mark = triangle*, color = magenta, dashed,error bars/.cd, y dir=both,y explicit] table[x=Minutes,y=PRIO PRESERVE, col sep=comma, y error = PRIO PRESERVE ERROR]{anc/Results-Strings_Min.csv};
\addplot+[mark = diamond*, color = teal, dashed,error bars/.cd, y dir=both,y explicit] table[x=Minutes,y=PRIO CONS PRESERVE, col sep=comma, y error = PRIO CONS PRESERVE ERROR]{anc/Results-Strings_Min.csv};
\end{axis}
\end{tikzpicture}
    \caption{Graph showing the effects of priority ordering for string data (showing standard error)}
    \label{fig:q3-string-results}
\end{figure}
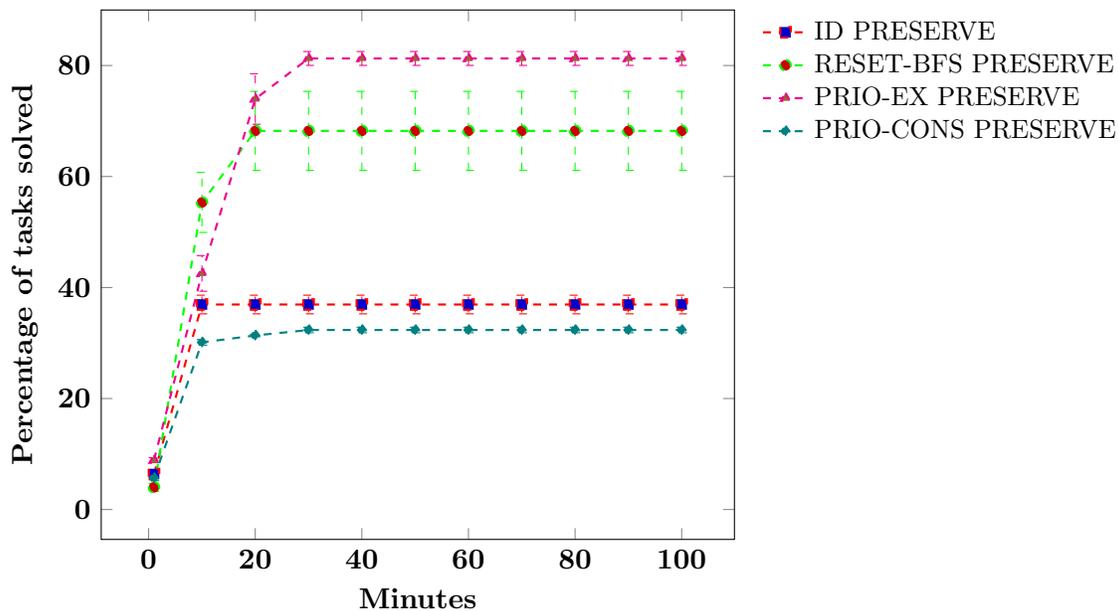

In \autoref{fig:q3-string-results} we observe the effect of using a priority ordering for the breadth first search approach. We have also plotted the performance of the state-of-the-art system to reference as a benchmark. The PRIO-EX algorithm, which sorts by the precision of the best program found until now, outperforms the original reset approach. On the other hand, the PRIO-CONS algorithm, which sorts by the number of constraints, exhibits below par performance. We associate this with the fact that the PRIO-CONS algorithm uses a poor heuristic. Tasks for which we have accumulated more constraints have a smaller hypothesis space size, but it is not a good indicator of being closer to a solution. 
Furthermore, by analysing \autoref{tab:string-table} we observe that priority ordering results in a lower variance in the performance of the system. We consider this a possible upside, as it improves the system's reliability. 
We conclude that the answer for \textbf{Q3} is \textbf{yes}, but we outline the need for further research in this area, with the purpose of developing more accurate heuristics.
\subsection{Robot movement}
This data set contains single examples for each task because any positive examples generated would be part of the same equivalence class (e.g. moving two cells to the right is independent of starting position). Consequently, there would be no benefit in testing the PRIO-EX algorithm against this data set. Moreover, the previous two data sets have shown that the PRIO-CONS approach performs poorly in practice. Therefore, we decide to conclude our answer for \textbf{Q3} based on the previous data and only test our claims for \textbf{Q1} and \textbf{Q2} against this data set.
\subsubsection{Overview}
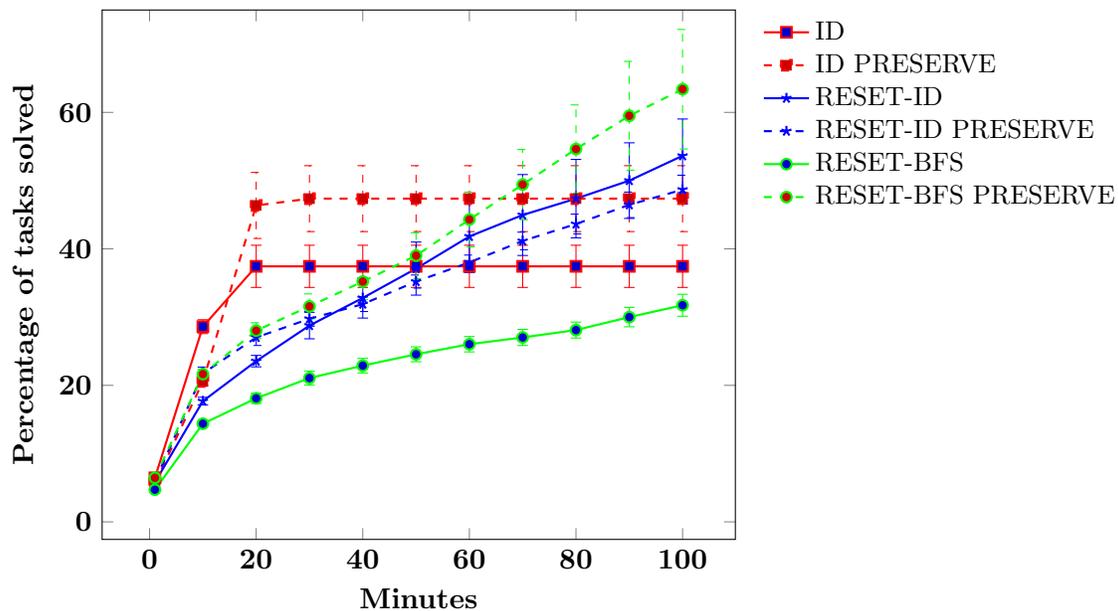
\begin{figure}[hbt!]
\centering
\begin{tikzpicture}
\begin{axis}[
    myplotstyle,
    legend pos=outer north east,
    legend entries={ID, ID PRESERVE, RESET-ID, RESET-ID PRESERVE, RESET-BFS, RESET-BFS PRESERVE},
    xlabel={Minutes},
    ylabel={Percentage of tasks solved},
    ymax = 75,
    width=10cm
]
\addplot+[mark = square*, color = red, solid,error bars/.cd, y dir=both,y explicit] table[x=Minutes,y=SOTA NO-PRESERVE, col sep=comma,y error = SOTA NO-PRESERVE ERROR]{anc/Results-Robots_Min.csv};
\addplot+[mark = square*, color = red, dashed,error bars/.cd, y dir=both,y explicit] table[x=Minutes,y=SOTA PRESERVE, col sep=comma, y error = SOTA PRESERVE ERROR]{anc/Results-Robots_Min.csv};
\addplot+[mark = star, color = blue, solid,error bars/.cd, y dir=both,y explicit] table[x=Minutes,y=RESET NO-PRESERVE, col sep=comma, y error = RESET NO-PRESERVE ERROR]{anc/Results-Robots_Min.csv};
\addplot+[mark = star, color = blue, dashed,error bars/.cd, y dir=both,y explicit] table[x=Minutes,y=RESET PRESERVE, col sep=comma, y error = RESET PRESERVE ERROR]{anc/Results-Robots_Min.csv};
\addplot+[mark = *, color = green, solid,error bars/.cd, y dir=both,y explicit] table[x=Minutes,y=RESET-IMP NO-PRESERVE, col sep=comma, y error = RESET-IMP NO-PRESERVE ERROR]{anc/Results-Robots_Min.csv};
\addplot+[mark = *, color = green, dashed,error bars/.cd, y dir=both,y explicit] table[x=Minutes,y=RESET-IMP PRESERVE, col sep=comma, y error = RESET-IMP PRESERVE ERROR]{anc/Results-Robots_Min.csv};
\end{axis}
\end{tikzpicture}
    \caption{Overview of robot results (showing standard error)}
    \label{fig:robot-results}
\end{figure}

In \autoref{fig:robot-results} we present an overview of the performance of each algorithm, expressed as the percentage of tasks solved after several minutes.
\subsubsection{Question 1}
By reviewing \autoref{fig:robot-results}, we observe that using constraint preservation has improved the performance of two out of three of the tested strategies. As expected, RESET-BFS benefits the most from constraint preservation. On the other hand, the RESET-ID approach performs slightly worse when constraint preservation is enabled. By manually analysing the results of the data, we associate this finding with noise generated by order randomisation. Therefore, the answer for \textbf{Q1} is \textbf{yes}.
\subsubsection{Question 2}
\begin{figure}[hbt!]
\centering
\begin{tikzpicture}
\begin{axis}[
    myplotstyle,
    legend pos=outer north east,
    legend entries={ID PRESERVE, RESET-ID PRESERVE, RESET-BFS PRESERVE},
    xlabel={Minutes},
    ylabel={Percentage of tasks solved},
    ymax = 75,
    width=10cm
]
\addplot+[mark = square*, color = red, dashed,error bars/.cd, y dir=both,y explicit] table[x=Minutes,y=SOTA PRESERVE, col sep=comma, y error = SOTA PRESERVE ERROR]{anc/Results-Robots_Min.csv};
\addplot+[mark = star, color = blue, dashed,error bars/.cd, y dir=both,y explicit] table[x=Minutes,y=RESET PRESERVE, col sep=comma, y error = RESET PRESERVE ERROR]{anc/Results-Robots_Min.csv};
\addplot+[mark = *, color = green, dashed,error bars/.cd, y dir=both,y explicit] table[x=Minutes,y=RESET-IMP PRESERVE, col sep=comma, y error = RESET-IMP PRESERVE ERROR]{anc/Results-Robots_Min.csv};
\end{axis}
\end{tikzpicture}
    \caption{Graph comparing state-of-the-art and reset strategies for robot data (showing standard error)}
    \label{fig:q2-robot-results}
\end{figure}
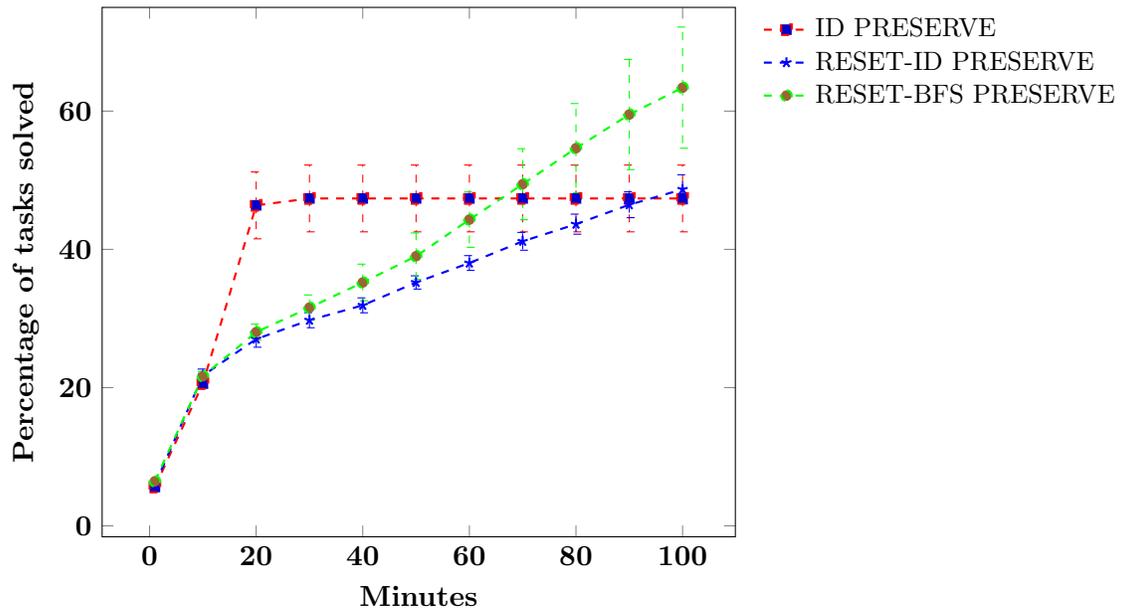

\autoref{fig:q2-robot-results} showcases a comparison between the state-of-the-art approach and the two reset strategies. The current state-of-the-art approach solves the first few tasks much quicker because it does not reset the hypothesis space size on each found solution. After the 20 minutes mark, we observe that the state-of-the-art approach stops making progress because it has reached a large enough search size to be able to find any more solutions in a reasonable time frame. Whereas both reset approaches show that they are making constant progress, with their learning rate remaining relatively constant\footnote{As mentioned in the introduction, we expect to observe a slight decay in learning rate as more tasks are added to the BK.}. Although the learning time of the first tasks is lower for the reset strategies, we evaluate that the ability to learn a higher percentage of tasks is more important when evaluating the performance of a system. Therefore, we conclude that the answer for \textbf{Q2} is \textbf{yes}.
\chapter{Conclusions}
\section{Summary}
Although an area that has shown high potential, there has been little development in terms of multi-task learning approaches in ILP. We have experimented with refining the current state-of-the-art approach with the expectation of obtaining substantially improved performance. We presented our assumptions' theoretical basis and then empirically confirmed them against three data sets. We have implemented both our new approach and the state-of-the-art technique based on the Popper ILP system. Our experimental results show that (1) constraint preservation improves learning speed in multi-task settings (\textbf{yes} to \textbf{Q1}), (2) the benefits of search size reset overcome it's disadvantages (\textbf{yes} to \textbf{Q2}), and (3) heuristic ordering of programs can improve learning rate (\textbf{yes} to \textbf{Q3}). 

In the following sections, we present the limitations of our work and suggest future work that could improve the performance of Popper in multi-task environments. 
\section{Limitations}
We have identified multiple factors that restricted the scope of our work.
\paragraph{Limited data.} The use of multi-task learning is relatively new to the field of ILP. Therefore, we highlight the lack of publicly available data sets used to benchmark such systems. We believe that our findings would have been better supported by testing in a wider variety of data, which have been previously found to uncover more use cases we might not have considered in our project.
\paragraph{Implementation improvement.} Our implementation of multi-task Popper has one specific drawback we have encountered during testing. On each reset operation, the underlying ILP environment is re-initialised, which often results in long reset times. There exists a simple solution to this issue, which would enforce the system always to preserve constraints between runs or to iterate through the constraint list and ``forget" each of them one by one. Such an approach would have introduced bias in our experimental results by favouring constraint preservation. We argue that a more specialised implementation of Popper would circumvent this problem and further improve learning time (especially for the reset approach, where this part of the code is called more frequently).
\paragraph{Bias settings.} One significant challenge in ILP is selecting bias parameters \cite{ade1995declarative}. Examples of bias parameters are the maximum number of variables allowed in a program or the maximum number of clauses allowed in a program. Appropriate selection of bias parameters has a significant impact on the learning rate by modifying the size of the possible hypothesis space. We argue that in the case of multi-task learning, the initial bias, which was set by the user with the initial BK in mind, may no longer be appropriate. For example, some tasks may have three clause solutions when solved by themselves, but only two clause solutions in the multi-task learning environment. Therefore, if our system could automatically compute bias parameters, it would improve the learning rate by constraining the hypothesis spaces size when the BK is augmented.
\section{Future work}
\paragraph{Explainable results.} Without multi-task learning, each program learnt is built from primitives, which are given in the BK and known to a human operator. When multi-task learning is introduced, program definitions may also contain other programs. This affects the ability of a human to understand the programs in two ways. Firstly, we note that, in a multi-task learning setting, knowledge of all the new predicates that appear in a solution is required before reading it. Secondly, we outline the fact that it is a common occurrence for more complicated tasks to be selected in favour of simpler tasks when they cannot be differentiated. For example see \autoref{fig:explainable-anc/Results-example}.
\begin{figure}[hbt!]
\centering
\begin{minted}{prolog}
    f0(A,B) :- draw1(A,B)
    f0(A,B) :- move_right(A,B),f0(A,B)
    % When learning a task f1(A,B), such as f1(A,B) :- draw1(A,B)
    % A valid solution for f1(A,B) is:
    f1(A,B) :- f0(A,B) 
    % But we expect that a user would prefer the trivial solution:
    f1(A,B) :- draw1(A,B)
\end{minted}
\caption{Example of two different ways in which a program can be learnt}
\label{fig:explainable-anc/Results-example}
\end{figure}

In the above case, both \texttt{draw1} and \texttt{f0} appear to behave identically for the system, in the absence of negative examples. In consequence, we propose the development of a predicate translator. A predicate translator would take the output of a multi-task ILP system and transform it such that all predicates will only make use of definitions from the BK, and any clause that does not affect the correctness of the task will be pruned out. 
\paragraph{Task ordering.} In this project, we have shown the potential of task ordering in improving the learning capacity of multi-task ILP systems. We suggest further research in developing more accurate heuristics to predict which tasks are closer to being solved. One of the solutions we propose is the use of other Machine Learning techniques in order to identify more manageable tasks and give them priority. We also concede the fact that there are certain limitations to the applicability of such an approach, especially when there are very few examples given for each task. 
\paragraph{Parallel computation.} When searching for solutions of a specific size, one could modify the current approach that attempts to solve tasks sequentially to try and solve multiple tasks concurrently, in a bag of tasks manner. This method would allow one to vastly increase the learning rate of multi-task Popper on multi-core systems. Moreover, a significant bottleneck observed during all experiments is that the system got stuck searching for solutions to unsolvable tasks. This issue will be mitigated, as data sets would require as many high difficulty tasks as threads to congest the system. 
\newpage
\bibliographystyle{plainnat}
\bibliography{biblio.bib}
    \end{document}